\documentclass[journal]{IEEEtran}
\ifCLASSINFOpdf
  \usepackage[pdftex]{graphicx}
  \usepackage{caption}
\else
\fi
\usepackage{amsthm,amsmath,amssymb}
\usepackage{mathrsfs}
\usepackage{graphicx}
\usepackage{multirow}

\usepackage{cite}

\usepackage[hyphens]{url}
\begin{document}
%
\title{Perturbation Inactivation Based Adversarial Defense for Face Recognition}
%
%
%

\author{Min~Ren,
        Yuhao~Zhu,
        Yunlong~Wang*,
        Zhenan~Sun,~\IEEEmembership{Senior Member,~IEEE,}
\thanks{M. Ren is with the School of Artificial Intelligence, University of Chinese Academy of Sciences, Beijing 100049, China, and with the Center for Research on Intelligent Perception and Computing, National Laboratory of Pattern Recognition, Institute of Automation, Chinese Academy of Sciences, Beijing 100190, China. E-mail: min.ren@cripac.ia.ac.cn}
\thanks{Y. Zhu is with the Postgraduate Department, China Academy of Railway Sciences, Beijing 100081, China. E-mail: zhuyuhao1994@outlook.com}
\thanks{Y. Wang is with the Center for Research on Intelligent Perception and Computing, National Laboratory of Pattern Recognition, Institute of Automation, Chinese Academy of Sciences, Beijing 100190, China. E-mail: yunlong.wang@cripac.ia.ac.cn}
\thanks{Z. Sun is with the Center for Research on Intelligent Perception and Computing, National Laboratory of Pattern Recognition, Institute of Automation, Chinese Academy of Sciences, Beijing 100190, China, and with CAS Center for Excellence in Brain Science and Intelligence Technology School of Artificial Intelligence, University of Chinese Academy of Sciences, Beijing 100190, China. E-mail: znsun@nlpr.ia.ac.cn}
\thanks{*Corresponding author: Yunlong Wang.}}

%
%

\markboth{IEEE Transactions on Information Forensics \& Security}%
{IEEE Transactions on Information Forensics \& Security}
%



\maketitle

\begin{abstract}
Deep learning-based face recognition models are vulnerable to adversarial attacks.
To curb these attacks, most defense methods aim to improve the robustness of recognition models against adversarial perturbations.
However, the generalization capacities of these methods are quite limited.
 In practice, they are still vulnerable to unseen adversarial attacks.
Deep learning models are fairly robust to general perturbations, such as Gaussian noises.
A straightforward approach is to inactivate the adversarial perturbations so that they can be easily handled as general perturbations.
In this paper, a plug-and-play adversarial defense method, named perturbation inactivation (PIN), is proposed to inactivate adversarial perturbations for adversarial defense.
We discover that the perturbations in different subspaces have different influences on the recognition model. 
There should be a subspace, called the immune space, in which the perturbations have fewer adverse impacts on the recognition model than in other subspaces.
Hence, our method estimates the immune space and inactivates the adversarial perturbations by restricting them to this subspace.
The proposed method can be generalized to unseen adversarial perturbations since it does not rely on a specific kind of adversarial attack method.
This approach not only outperforms several state-of-the-art adversarial defense methods but also demonstrates a superior generalization capacity through exhaustive experiments.
Moreover, the proposed method can be successfully applied to four commercial APIs without additional training, indicating that it can be easily generalized to existing face recognition systems.
The source code is available at \url{https://github.com/RenMin1991/Perturbation-Inactivate}.
\end{abstract}


%
\IEEEpeerreviewmaketitle

%
%
%
%


\section{Introduction}
\label{Introduction}

\IEEEPARstart{D}{eep} learning-based feature extractors have achieved great success in various fields, including image classification~\cite{L1998Gradient, Krizhevsky2012ImageNet, Simonyan2014Very, Szegedy2015Going, Huang2017Densely, Hu2018Squeeze, He2016Deep}, object detection~\cite{girshickICCV15fastrcnn, Redmon2015You}, and semantic segmentation~\cite{Jonathan2014Fully, He2017Mask}, due to their capability to perform nonlinear mappings from raw data to high-level features.
While CNNs are powerful, they are also vulnerable to adversarial examples generated by various attacking algorithms~\cite{Szegedy2013Intriguing, Goodfellow2014Explaining, S2017Universal}. 
Therefore, many  researchers have focused on the design of adversarial defense methods for deep learning models to resist attackers~\cite{Kurakin2016Adversarial, Na2018Cascade, Tram2018Ensemble, Ross2017Improving}.

As one of the most ubiquitous applications of deep learning technology,  CNN-based face recognition systems~\cite{Taigman2014DeepFace, Yi2014Deep, Schroff2015FaceNet, Liu2017SphereFace, Wang2018CosFace, Deng2018ArcFace} can outperform human beings in both verification and identification scenarios.
Face recognition technology has been widely used in various fields of identity authentication, where security and robustness are essential for the system.
However, CNN-based face recognition algorithms are also vulnerable to adversarial attacks from cyberspace~\cite{Dong2019Efficient} or in the real world~\cite{Komkov2013AdvHat}.
Hence, defense against adversarial attacks for face recognition is an essential research topic.

\begin{figure}[t]
\begin{center}
\setlength{\abovecaptionskip}{0pt}
\setlength{\belowcaptionskip}{0pt}
\includegraphics[width=\linewidth]{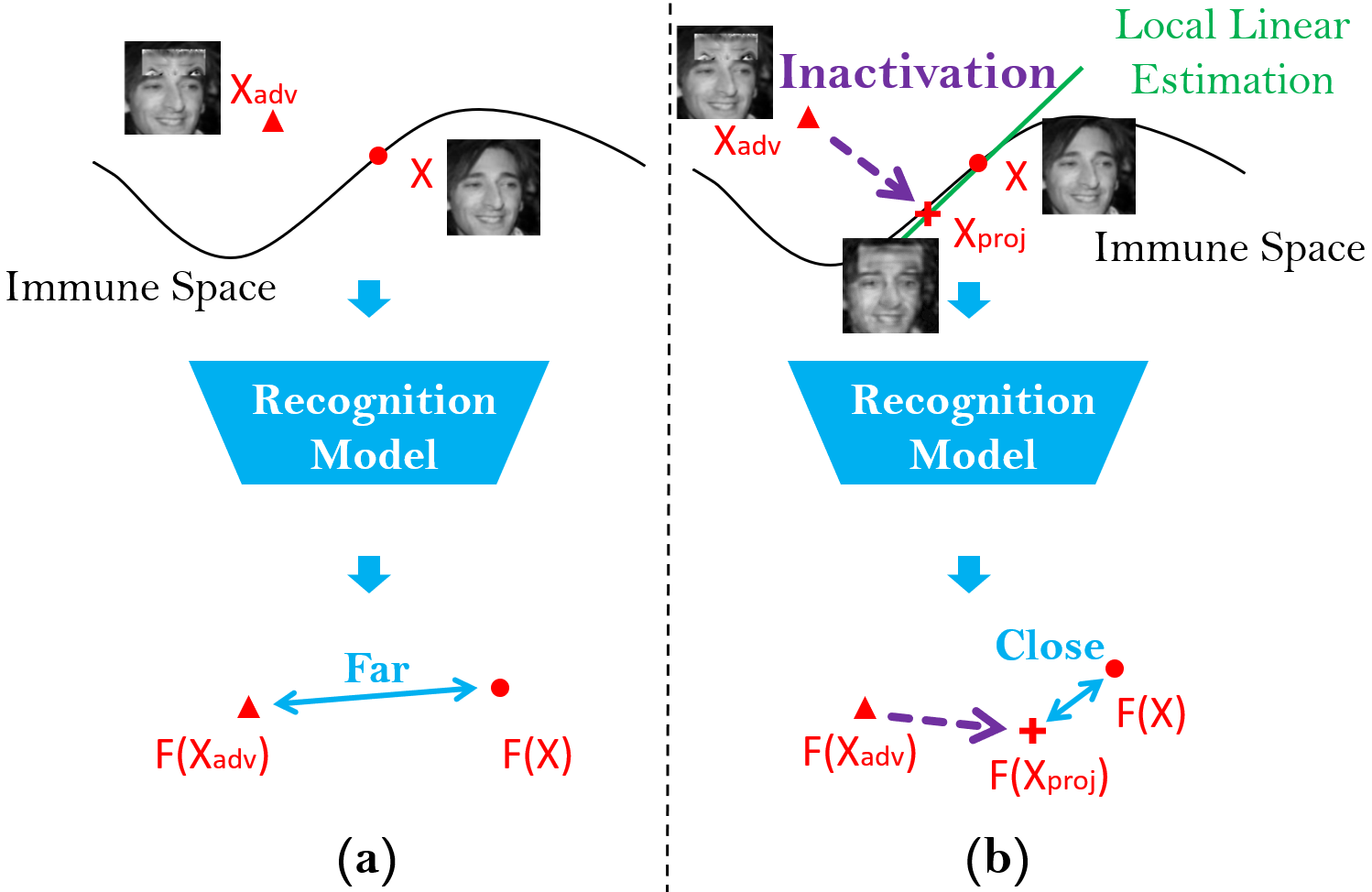}
\end{center}
\setlength{\abovecaptionskip}{0pt}
\setlength{\belowcaptionskip}{0pt}
   \caption{Schematic diagram of the proposed adversarial defense framework. \textbf{(a)} $X_{adv}$ is an adversarial sample generated from X by sticker attack~\cite{Komkov2013AdvHat}. $F(\cdot)$ is the recognition model, which is a function that maps input samples into features. \textbf{(b)} $X_{adv}$ is projected into the estimation of the immune space to inactivate the adversarial perturbation before the recognition model.}
\label{fig:Idea}
\end{figure}

There are two challenges to the adversarial defense of face recognition.
First, we cannot know the attack method in advance.
In real-world applications, the face recognition system is exposed to various possible attacks. Considering all of these attacks are intractable. Furthermore, new attack methods are constantly emerging.
%
Hence, the generalization capacity to unseen adversarial attacks is crucial for a defense method for face recognition.
Most of the current adversarial defense methods aim to improve the robustness of recognition models by adversarial training or a similar strategy, which means including adversarial samples in the training of recognition models.
However, the generalization capacity of adversarial training-based methods cannot be guaranteed, particularly for unseen attacks.
%


Second, the attackers are aware of the in-service defender in practice.
Once the defense method has been deployed in the face recognition system, the defense method itself also becomes the target of attackers.
This scenario is called \emph{online defense} in our paper, and we will formally define it in Sec.~\ref{Experiment}.
It is more difficult for the defense method to protect the recognition model.
Most current defense methods are not designed for this tough scenario.


In fact, the deep learning models are fairly robust to general perturbations~\cite{Szegedy2013Intriguing}, such as Gaussian noise, even though they are vulnerable to adversarial perturbations. 
Similarly, the human body is immune to most viruses, except a few kinds of infectious virus. 
Rather than expecting human beings to evolve to be immune to all kinds of viruses, which involves great uncertainty and is extremely costly, a more practical approach to resist these powerful viruses is  to inactivate them until they can be easily treated as general viruses. 
By analogy, a proper scheme for face recognition adversarial defense is inactivating the adversarial perturbations until they are mediocre as general noises, rather than improving the immunity of recognition models, which is much more costly and lacks generalization capacity.
This strategy leverages the inherent immunity of the recognition model to perturbations. Adversarial samples are not involved during training.
Hence, this approach will not overfit to a specific attack method.

To stimulate the recognition model's inherent immunity, we explore the robustness of CNN-based face recognition models under different kinds of perturbations.
We find that the subspace of perturbations is a key factor of the effect on the recognition models.
Perturbations in different subspaces have different effects on the similarity metrics of recognition models.
Accordingly, we hypothesize that there exists a subspace in which the perturbations have fewer adverse impacts on the similarity metrics than in other subspaces.
This subspace is named the immune space in this paper.
Given the immune space, an adversarial sample can be projected into it to inactivate the adversarial perturbation.
%
Hence, the adversarial perturbations will be restrained in the subspace where the perturbations have the least adverse impacts.
Note that clean samples are in the immune space by default.
In \emph{online defense}, the search space of attack methods will be significantly compressed, and it will be much more difficult to dupe the recognition model.

To estimate the immune space, we propose a novel framework in this paper.
Inspired by principal component analysis (PCA), which is a classical method for linear subspace estimation, we propose a learnable PCA in this paper to estimate the immune space.
In classical PCA, the estimated subspace is expanded by eigenvectors that are selected according to eigenvalues.
The samples are projected in this subspace for reconstruction.
Considering that the immune space can be highly nonlinear, the linear estimation provided by classical PCA is not appropriate.
To estimate the immune space properly, the proposed learnable PCA aims to provide a unique estimation for each sample.
The estimation for each sample is a linear subspace in the vicinity of this sample, as shown in Fig. \ref{fig:Idea}, which means that the proposed learnable PCA estimates the immune space by a piecewise-linear strategy.
The estimation for each sample is also expanded by eigenvectors, and the eigenvectors are selected by an agent in the form of a deep neural network, rather than according to eigenvalues.
Deep reinforcement learning is adopted to train the proposed learnable PCA.



The main contributions of this paper can be summarized as follows:

\begin{itemize}

\item We explore the robustness of the face recognition model under different kinds of perturbations. 
It is found that the subspace of perturbations is a key factor determining the impact on recognition models. 

\item Based on this discovery, we propose an adversarial defense scheme that inactivates the adversarial perturbations by projecting the adversarial samples into the immune space.

\item  To estimate the immune space, we propose a novel learnable PCA as the key component of the proposed adversarial defense method. The learnable PCA provides a piecewise-linear estimation of the immune space.

\end{itemize}

The remainder of this paper is organized as follows:
Section~\ref{RelateWork} presents a brief literature review of the related work.
The exploration of the robustness of the face recognition model under different kinds of perturbations is presented in Section~\ref{DiffPert}. 
The proposed defense method is described in detail in Section~\ref{Method}. 
The configurations and results of experiments are presented in Section~\ref{Experiment}.
Finally, the conclusions of this paper are summarized in Section~\ref{Conclusion}.



\section{Related Work}
\label{RelateWork}

\textbf{Deep learning based face recognition:}
Face recognition based on deep learning has achieved great success in recent years.
The first method for  CNN-based face representation was proposed by Taigman et al.~\cite{Taigman2014DeepFace}. A framework employing multiple convolutional networks was proposed by Yi et al.~\cite{Yi2014Deep}. A triplet loss function was applied to obtain a 128-D face embedding representation in~\cite{Schroff2015FaceNet}. A light convolutional architecture was proposed for face recognition in~\cite{wu2018light}. Recently, numerous methods for better feature embedding have been proposed, including SphereFace~\cite{Liu2017SphereFace}, CosFace~\cite{Wang2018CosFace}, ArcFace~\cite{Deng2018ArcFace} and so on.

Despite these significant advances, the deep learning models for face recognition are vulnerable to adversarial attacks~\cite{Dong2019Efficient, Komkov2013AdvHat, Zhong2020Towards}, which poses a serious threat to the security of face recognition systems.


\textbf{Adversarial attack:}
The adversarial attacking method for computer vision tasks is a research hotspot. Szegedy et al.~\cite{Szegedy2013Intriguing} first demonstrated that deep neural networks are vulnerable to adversarial perturbations. Subsequently, many adversarial attacking methods have been proposed. Goodfellow et al.~\cite{Goodfellow2014Explaining} proposed an efficient single-step attacking method named FGSM, which is a gradient-based method. DeepFool~\cite{Moosavi2016DeepFool} seeks to find the nearest decision boundary to confuse the model. C\&W~\cite{Nicholas2016Towards} was proposed to solve the joint optimization of objective function and scale of perturbation. Projected gradient descent (PGD)~\cite{Madry2017Towards} iteratively applies FGSM. Su et al.~\cite{Su2017One} proposed an interesting method to confuse the deep learning model by only changing a single pixel in the image. Generalization of the adversarial perturbations has also been reported. Attacking methods based on universal adversarial perturbation are proposed in~\cite{S2017Universal, Zhang2021Data, Wang2021Feature, Yuan2021Meta}.

Adversarial attacking methods tailored for face recognition were also reported recently. Dong et al.~\cite{Dong2019Efficient} proposed a decision-based black-box attacking method for face recognition.
Adversarial attacks in the real world are also reported in the literature.
Sharif et al.~\cite{Sharif2016Accessorize} proposed a systematic method to automatically generate physically realizable attacks through printing a pair of eyeglass frames.
Sticker attack is proposed in~\cite{Komkov2013AdvHat}, which attacks the face recognition system with a specifically designed rectangular paper sticker.
The real-world attacks set up new challenges to the defense approaches for face recognition systems. 


\textbf{Adversarial defense:}
The existing adversarial defense methods can be roughly classified into two categories. 
Methods in the first category aim to improve the robustness of neural networks against adversarial examples. 
Methods in the second category attempt to erase the adversarial perturbations from the samples before feeding them to the target model.
These two categories of methods will be reviewed separately.

A common strategy of the first kind is training the networks with adversarial examples~\cite{Kurakin2016Adversarial, Goodfellow2014Explaining, Na2018Cascade, Tram2018Ensemble}. 
This is a straightforward approach to improve the robustness against adversarial attack. 
Various learning strategies have been proposed to improve the robustness against gradient-based attacks. 
Ross et al.~\cite{Ross2017Improving} trained the model while regularizing input gradients. 
Network distillation~\cite{Papernot2015Distillation}, region-based classifier~\cite{Cao2017Mitigating}, generative model~\cite{Lee2017Generative, Jang2019ICCV}, and self-supervised learning~\cite{Moayeri2021Sample} are also adopted to improve the robustness of the models.
Rakin et al.~\cite{Rakin2018Parametric} proposed a trainable randomness method to improve the robustness by adversarial training.
Novel loss functions are proposed for adversarial defense in~\cite{Chen2019ICCV}. Mustafa et al.~\cite{Mustafa2019ICCV} improved the robustness by restricting the hidden space of deep neural networks. Zhong et al.~\cite{Zhong2019adversarial} adopted margin-based triplet embedding regularization to train the recognition model.
Cazenavette et al.~\cite{Cazenavette2021Architectural} attempt to improve the adversarial robustness of CNN by reframing each layer as a sparse coding model.
However, the generalization ability of these methods to unseen attacks cannot be guaranteed, and they are costly in practice.
The results of our experiments show that their performance on unseen attacks is not promising.

Some other methods are designed to remove the adversarial perturbations before the recognition model. 
Das et al.~\cite{Das2017Keeping} sought to remove the perturbations by JPEG compression. 
Image quilting and total variation minimization (TVM) are evaluated in~\cite{Guo2017Countering}. 
Meng et al.~\cite{Meng2017MagNet} proposed a two-pronged defense approach to remove the adversarial perturbations. 
Liao et al.~\cite{Liao2018Defense} adopt the U-Net~\cite{ronneberger2015Unet} as a denoising module to remove the adversarial perturbations.
PixelCNN~\cite{Salimans2017PixelCNN} is used to transform the adversarial examples to clean images in~\cite{Song2018PixelDefend}. Hilbert scan is applied to PixelCNN to improve the defense performance in~\cite{Bai2019ICCV}. 
Dezfooli et al.~\cite{Moosavi2018Divide} and Sun et al.~\cite{Sun2018Adversarial} adopted sparse coding to reconstruct patches of images. 
Gupta et al.~\cite{Gupta2019ICCV} attempted to find the most influential parts of the image for reconstruction.
Xie et al.~\cite{xie2019feature} adopt a self-attention layer to recover the original information in feature space.
Self-supervised learning is adopted to remove the adversarial noise in class activation feature space~\cite{Zhou2021Removing}.
However, most of this kind of methods are designed for general image classification.
Face  recognition is commonly a full or half open-set problem, which means that there is partial or no identity overlap between the training and test datasets.
The similarity metrics of the samples rather than the classification boundary are the criteria for recognition.
Hence, most of the defense method for general image classification are not suitable for face recognition.



\section{Robustness Against Perturbations in Different Subspaces}
\label{DiffPert}

To arouse the inherent robustness of recognition models against perturbations, we explore the robustness of the face recognition model under different kinds of noises. 
It is found that noises in different subspaces have significantly different effects on the similarity metrics between the samples.

\begin{figure}[t]
\begin{center}
\setlength{\abovecaptionskip}{0pt}
\setlength{\belowcaptionskip}{0pt}
\includegraphics[width=\linewidth]{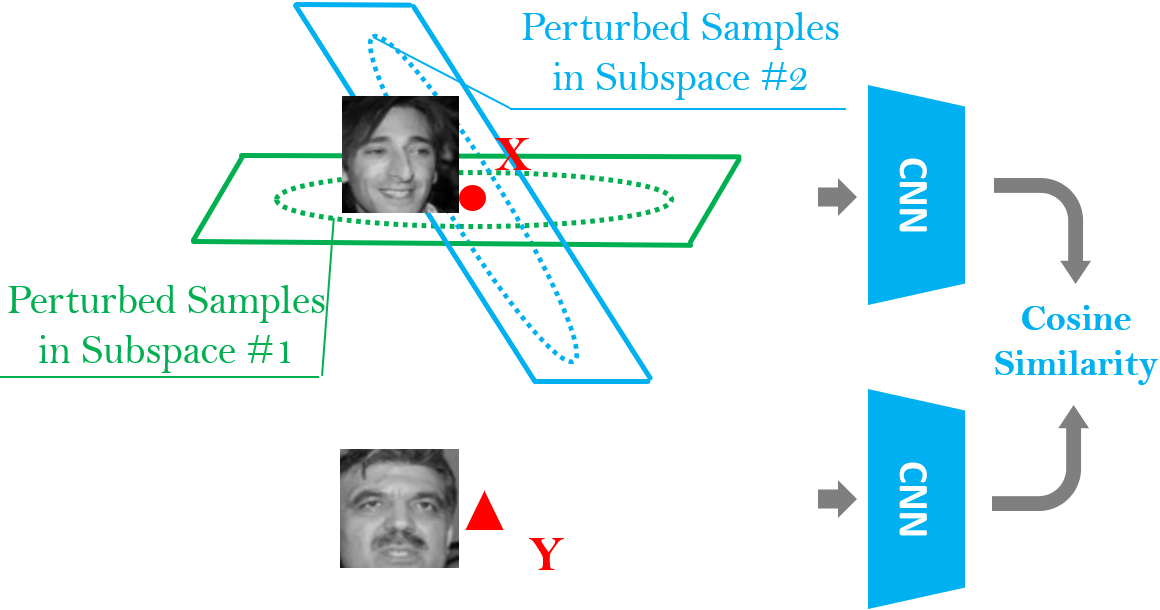}
\end{center}
\setlength{\abovecaptionskip}{0pt}
\setlength{\belowcaptionskip}{0pt}
   \caption{An image pair $X, Y$, $X$ is perturbed by Gaussian noises in different subspaces while $Y$ stays clean. Then, their similarity is computed by the recognition model. The perturbed samples in different subspaces have different effects on the distance metrics.}
\label{fig:Gauss_noise}
\end{figure}

ArcFace~\cite{Deng2018ArcFace}, which is a state-of-the-art face recognition model, is adopted as the feature extractor in this section. 
Two pairs of samples from Labeled Faces in the Wild (LFW)~\cite{Learned2016Labeled}  are selected randomly in this experiment. 
The first pair is a positive pair, and the second is a negative pair.
For each image pair, one of the images is contaminated by Gaussian noises, while the other remains clean. 
Cosine similarity of the features is used to measure the behavior of the recognition model.
%
%

\begin{figure}[h]
\begin{center}
\setlength{\abovecaptionskip}{0pt}
\setlength{\belowcaptionskip}{0pt}
\includegraphics[width=\linewidth]{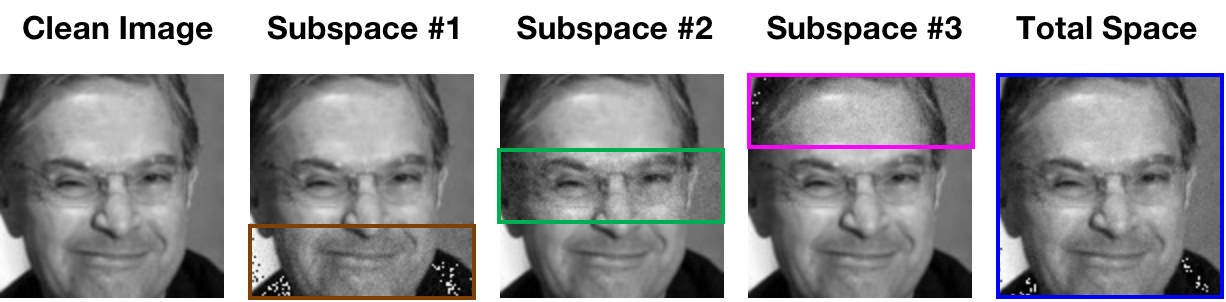}
\end{center}
\setlength{\abovecaptionskip}{0pt}
\setlength{\belowcaptionskip}{0pt}
   \caption{The three subspaces are the sets of pixels in three subregions: the bottom, middle, and top regions of the face image. The total space is also explored for comparison.}
\label{fig:ToySubspace}
\end{figure}

The total space of images is $\mathbb{R}^{H\times W}$, where $H$ and $W$ are the width and height of the images, respectively.
To explore the influence of noises in different subspaces, noises in three different subspaces are taken into consideration in this experiment: $\mathbb{R}_1^D$, $\mathbb{R}_2^D$, and $\mathbb{R}_3^D$, where $D$ is the dimension of this subspace, as shown in Fig.~\ref{fig:Gauss_noise}. 
The three subspaces are the sets of pixels in three subregions: bottom, middle, and top region of the face image, as shown in Fig.~\ref{fig:ToySubspace}.
The total space is also explored for comparison.
The intensity (L2 norm) of the noises is fixed to 0.04, and 10,000 perturbed samples are generated in each subspace.

\begin{figure}[h]
\begin{center}
\setlength{\abovecaptionskip}{0pt}
\setlength{\belowcaptionskip}{0pt}
\includegraphics[width=0.8\linewidth]{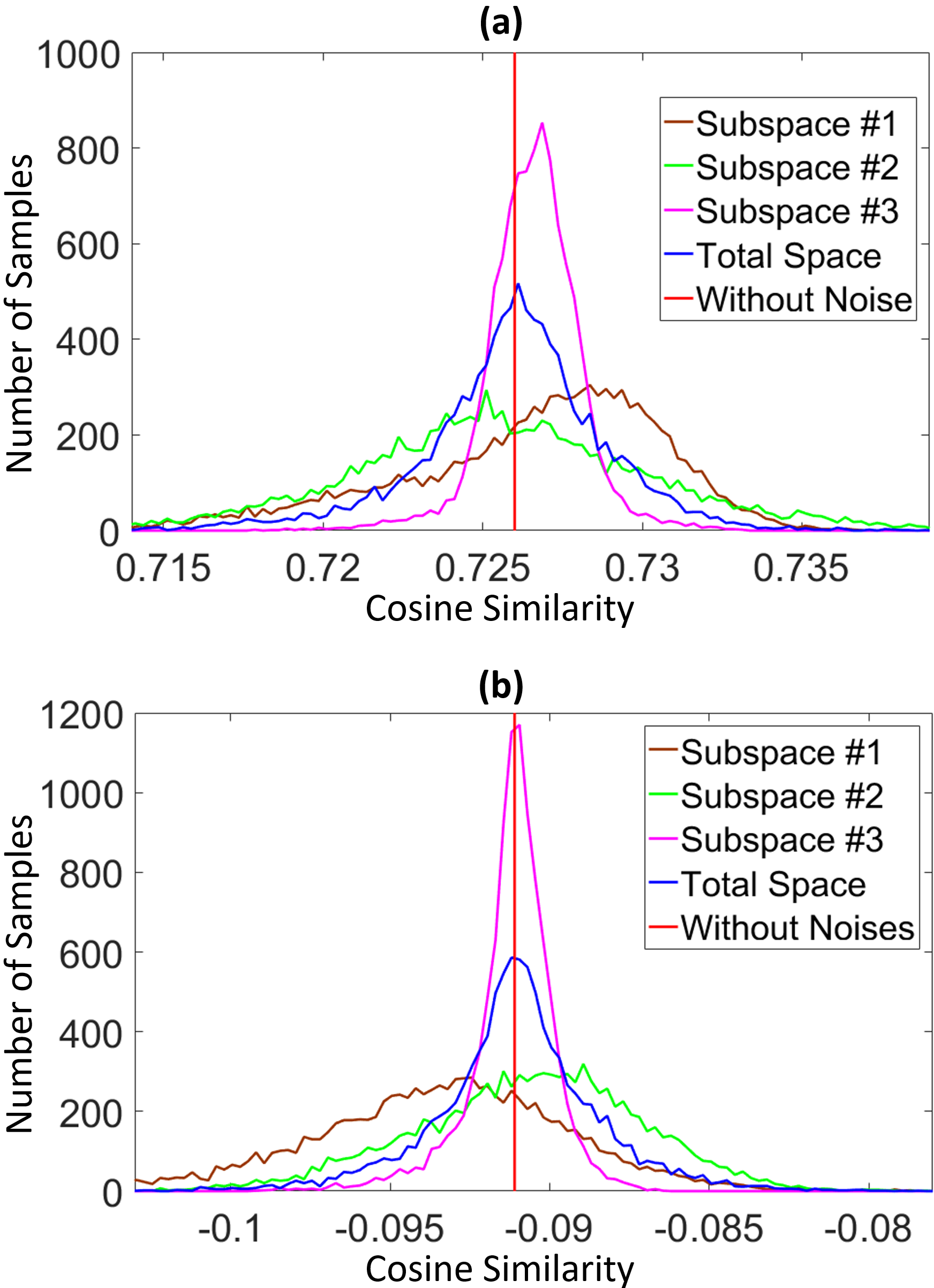}
\end{center}
\setlength{\abovecaptionskip}{0pt}
\setlength{\belowcaptionskip}{0pt}
   \caption{\textbf{(a)} Distribution of similarities of the positive pair. \textbf{(b)} Distribution of similarities of the negative pair. The effects of noises in different subspaces are different.}
\label{fig:SimDis}
\end{figure}

The distribution of similarities of each subspace is shown in Fig.~\ref{fig:SimDis}.
An examination of the distributions shows that the effects of the noises in different subspaces are quite different. 
The peaks of the distributions of subspaces \#1 and \#2 are in opposite directions compared to the similarity of the clean image. 
Meanwhile, the peak of the distribution of subspace \#3 remains close to that of the clean image.
Additionally, the variances of the distributions vary dramatically. 
The distribution of subspace \#3 is significantly more compact than the distributions of subspaces \#1 and \#2, indicating that the robustness of the recognition model against noises in subspace \#3 is better than that in the other two subspaces.

These distributions show that the robustness against noises of the recognition model varies from subspace to subspace, and this phenomenon can also be observed in other image pairs. 
The reason is that the significance of the information in the three subspaces is different.
The information in subspaces \#1 and \#2 is more important than the information in subspace \#3 for recognition. 
Hence, the recognition model is more robust to the noises in subspace \#3.
Hence, we hypothesize that there exists an immune space in which the perturbations have fewer impacts on the distance metrics than other subspaces, and the key to arousing the inherent robustness of recognition models is to find the immune space so that the perturbations can be inactivated by projecting the adversarial samples into this space.

\begin{figure}[h]
\begin{center}
\setlength{\abovecaptionskip}{0pt}
\setlength{\belowcaptionskip}{0pt}
\includegraphics[width=\linewidth]{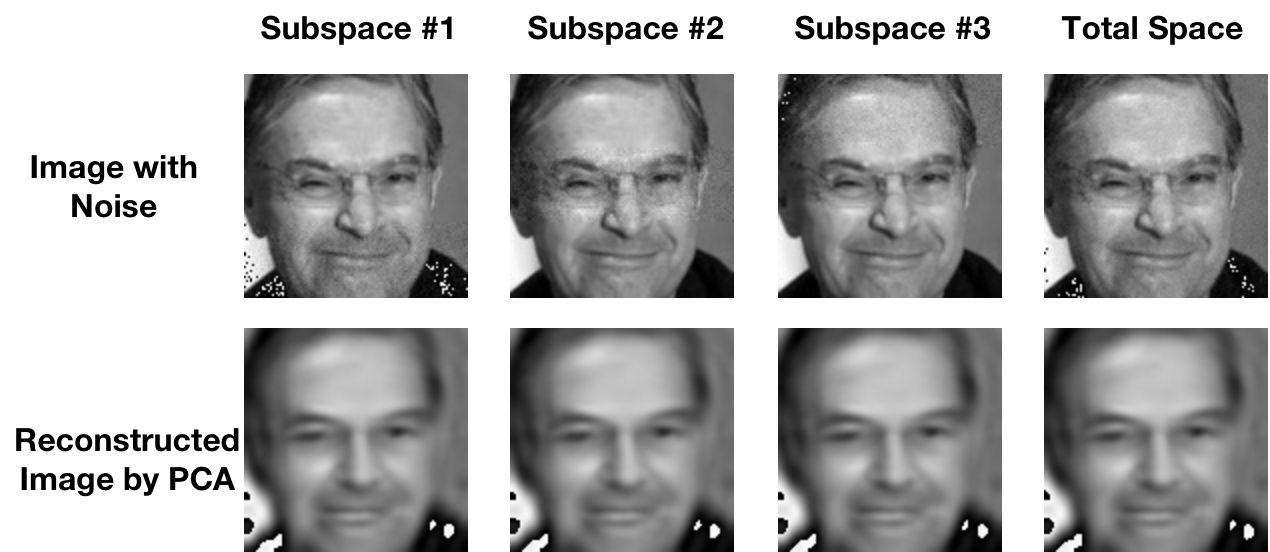}
\end{center}
\setlength{\abovecaptionskip}{0pt}
\setlength{\belowcaptionskip}{0pt}
   \caption{Examples of the reconstructed images by projection. The differences brought by noises are reduced by projection. But some of the useful information is lost at the same time.}
\label{fig:ToyRecon}
\end{figure}

\begin{figure}[h]
\begin{center}
\setlength{\abovecaptionskip}{0pt}
\setlength{\belowcaptionskip}{0pt}
\includegraphics[width=0.8\linewidth]{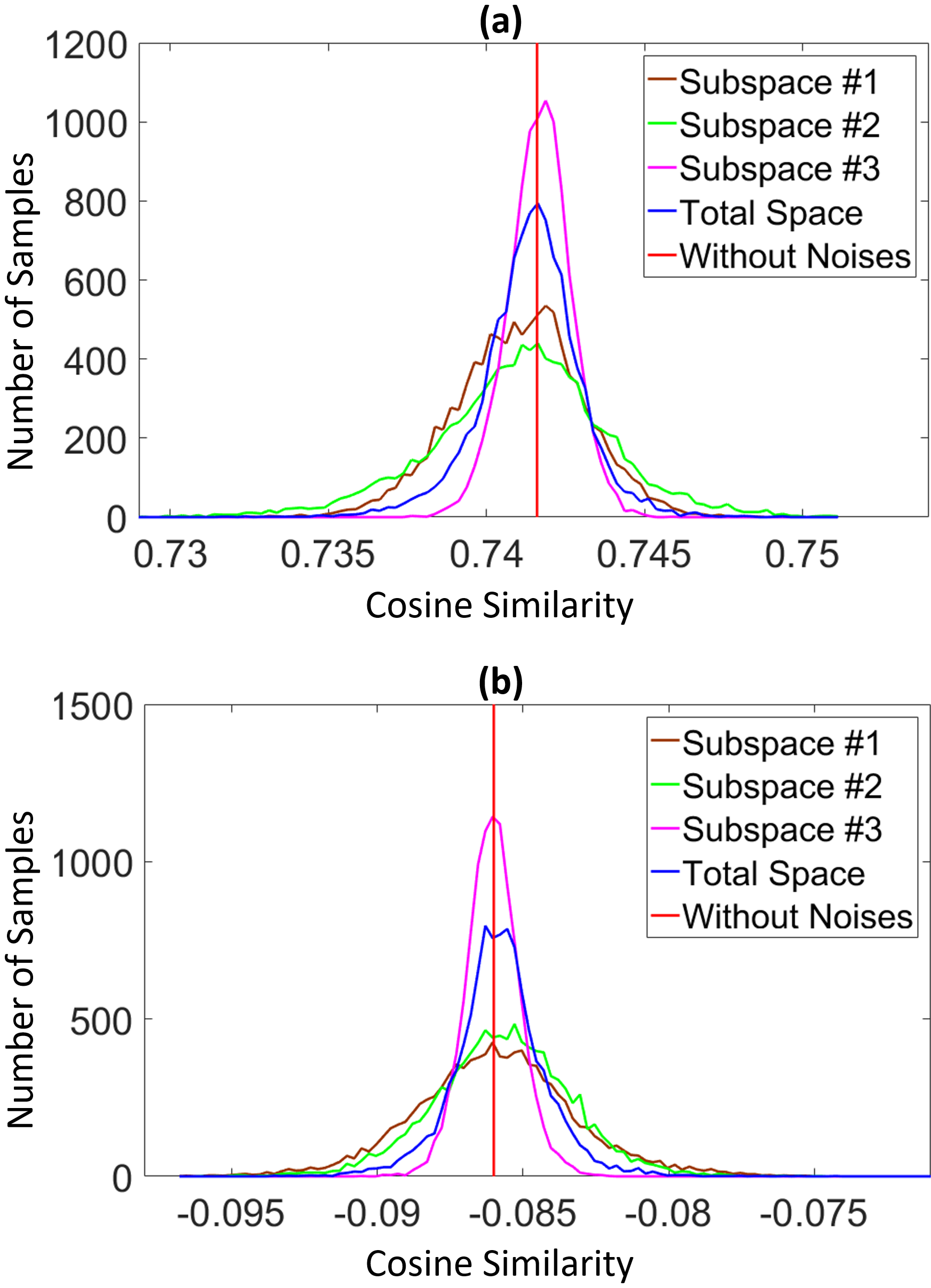}
\end{center}
\setlength{\abovecaptionskip}{0pt}
\setlength{\belowcaptionskip}{0pt}
   \caption{\textbf{(a)} Distribution of similarities of the positive pair after inactivation. \textbf{(b)} Distribution of similarities of the negative pair after inactivation. The adverse effects of the noises in all subspaces are reduced after inactivation.}
\label{fig:SimDisRecon}
\end{figure}

A toy experiment is conducted to verify this idea in this section. 
Principle component analysis (PCA) is a classical method for linear subspace learning and estimates the subspace through the first few eigenvectors. 
%
Accordingly, PCA is adopted to provide the estimation of the immune space for perturbation inactivation.
In this experiment, the first 200 eigenvectors are selected to expand the subspace into which the perturbed samples are projected. 
All of the perturbed images mentioned above are projected into this subspace for inactivation (the clean image is also reconstructed for comparison).
Examples of the projected images by projection are shown in Fig.~\ref{fig:ToyRecon}.
The similarity distribution after inactivation is shown in Fig.~\ref{fig:SimDisRecon}.

The peaks of all of the distributions are pulled back to the similarity between the clean images, as shown in the figures. The variances of all of the distributions are decreased.
The similarity distributions of subspaces \#1 and \#2 are affected more significantly because eigenvectors with higher eigenvalues are more likely to contain the useful information of the face image, and the noises in these two subspaces are more probably orthogonal to these eigenvectors.
These noises are more probable to distribute in the space of the eigenvectors with lower eigenvalues so that they can be removed by projection.
This means that the adverse effects of the noises are reduced by inactivation.

\begin{figure}[h]
\begin{center}
\setlength{\abovecaptionskip}{0pt}
\setlength{\belowcaptionskip}{0pt}
\includegraphics[width=0.9\linewidth]{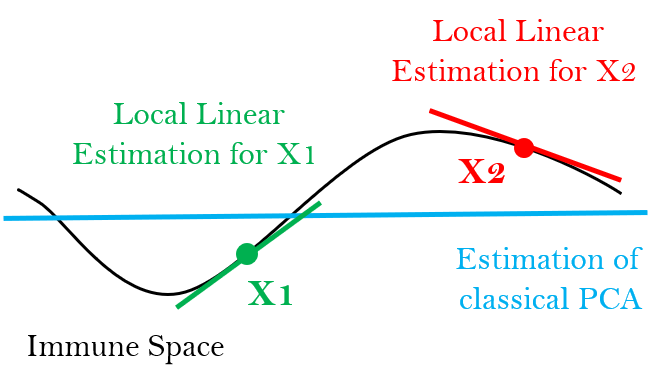}
\end{center}
\setlength{\abovecaptionskip}{0pt}
\setlength{\belowcaptionskip}{0pt}
   \caption{Classical PCA estimates the immune space linearly. However, the immune space can be highly nonlinear. The proposed learnable PCA piecewise-linearly estimates the immune space by local linear subspace.}
\label{fig:PCASpace}
\end{figure}

However, the similarities between the clean images are also changed, as observed from the figures, thereby degrading the face recognition performance.
The reason is that PCA estimates the immune space linearly, but the immune space can be highly nonlinear since the distribution of facial images in the real world is quite complex. 
The linear estimation conducted by PCA is too loose for perturbation inactivation. 
To solve this problem, the proposed learnable PCA provides a piecewise-linear estimation to maintain the distance metrics of the clean images well and simultaneously inactivate the adversarial perturbations.
It provides a local linear subspace nearby each sample, as shown in Fig.~\ref{fig:PCASpace}.
The nonlinear immune space can be well estimated by the set of local linear estimations.

\section{Perturbation Inactivation}
\label{Method}


\subsection{Overview}
\label{MetOver}

According to the analysis in Section~\ref{DiffPert}, the proposed method aims to estimate the immune space in a piecewise-linear manner.
Then, the perturbed image can be projected into the estimated subspace to inactivate the adversarial perturbations before recognition models.

\begin{figure}[t]
\begin{center}
\setlength{\abovecaptionskip}{0pt}
\setlength{\belowcaptionskip}{0pt}
\includegraphics[width=\linewidth]{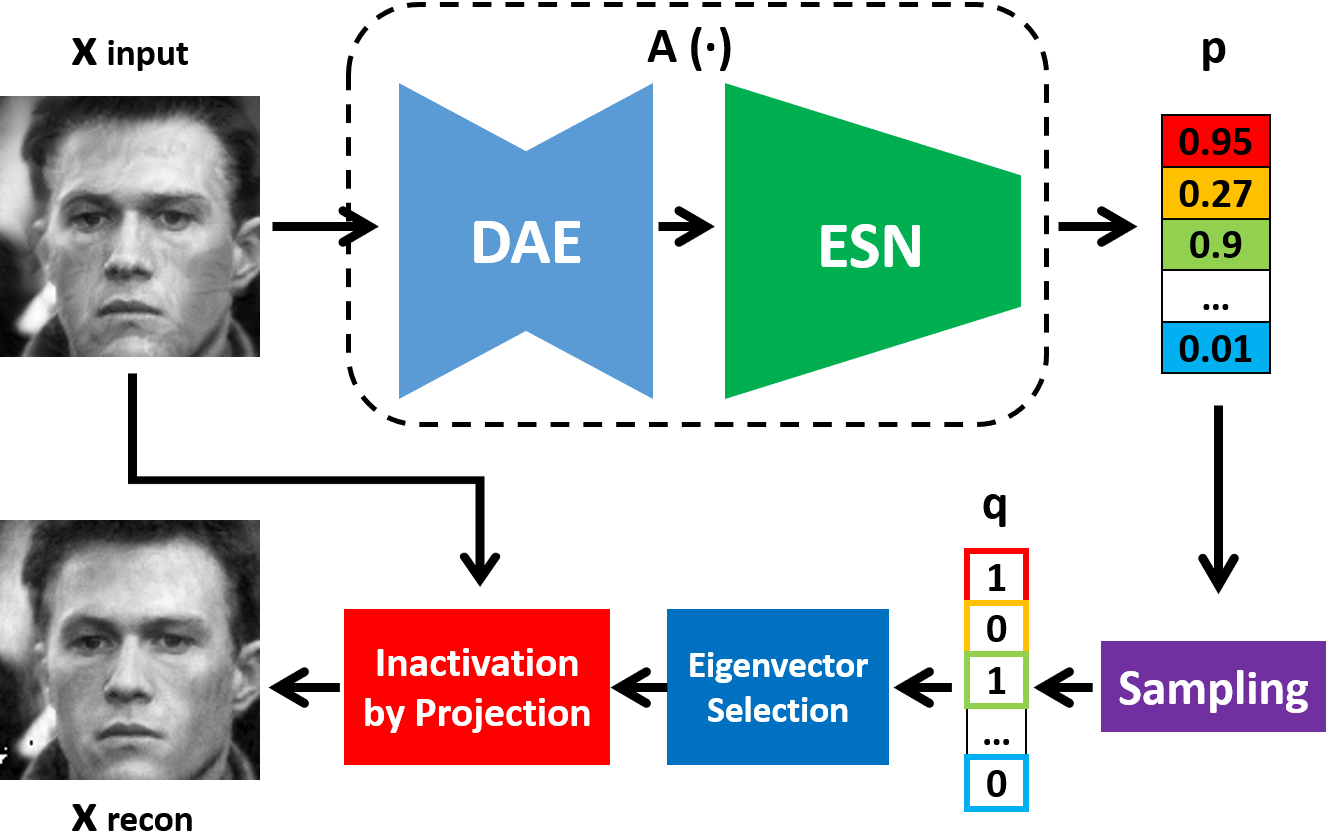}
\end{center}
\setlength{\abovecaptionskip}{0pt}
\setlength{\belowcaptionskip}{0pt}
   \caption{The perturbations of the input image are inactivated by projection into the subspace, which is an estimation of the immune space, expanded by the selected eigenvectors. The eigenvectors are selected according to the probabilities given by an agent $A(\cdot)$.}
\label{fig:TestPipline}
\end{figure}

Inspired by PCA, the proposed method estimates the immune space by selecting eigenvectors adaptively in the vicinity of each sample. 
The selection of eigenvectors is carried out by an agent in the form of a deep neural network. The input of the network is an image, and the output is the selection probabilities of the eigenvectors:
\begin{equation}
A_{\theta}(x) = p
\end{equation}
where $A_{\theta}(\cdot)$ is the agent, $p\in \mathbb{R}^N$ is the output vector, and $N$ is the number of eigenvectors. Each component of $p$ is a probability of selecting a specific eigenvector.
%
 %
By sampling according to all components of $p$, we can obtain a binary vector $q\in \{ 0, ~1 \}^N$, as shown in Fig.~\ref{fig:TestPipline}.
Eigenvectors are selected according to $q$ for estimation: 1 for selected, 0 for unselected. 
Then, the input image is projected into the subspace expanded by the selected eigenvectors to inactivate the adversarial perturbation. This mechanism is named learnable PCA.

The key difference between vanilla PCA and the learnable PCA is that rather than a linear estimation, the learnable PCA provides a piecewise-linear estimation of the immune space in the vicinity of each samples.

\subsection{Learnable PCA}
\label{MetLeaPCA}

The learnable PCA contains three components. The first component is a pre-trained denoising autoencoder (DAE), the second component is a convolutional neural network named eigenvector selection network (ESN), and the third component is two fully connected layers followed by sigmoid activation.

The pre-trained denoising autoencoder is used to accelerate the convergence of the training process and is frozen during the training of the other learnable PCA components. 
%
%
The architecture of the pretrained denoising autoencoder is shown in Tab.~\ref{tab:DAE}.
The eigenvector selection network (ESN) is a convolutional neural network that is adapted to extract deep representations of images. The ESN architecture is shown in Tab.~\ref{tab:EPN}.
The third component transforms the representations to an $N$-dimensional probability vector $p$. The architecture of the third component is shown in Tab.~\ref{tab:FC}.

The binary selection vector $q$ is generated by sampling $N$ Bernoulli distributions, which are parameterized by the components of the probability vector $p$.
According to the selection vector $q$, the matrix consisting of the selected eigenvectors $B_q \in \mathbb{R}^{D\times N}$ is generated, where $D$ is the dimension of the input image.
Then, the input image $x_{input}$ can be reconstructed by $B_q$:
\begin{equation}
x_{recon} = B_q B_q^T (x_{input}-x_m) +x_m
\end{equation}
where $x_m$ is the mean of all clean images in the training set.

\begin{table}[h]
\begin{center}
\caption{Architecture of the pretrained DAE}
\label{tab:DAE}
\setlength{\tabcolsep}{1mm}
{
\begin{tabular}{c|c|c|c}
\hline
\bf{Layer Name}&\bf{Output Size}&\bf{Kernel Size}&\bf{Stride}\\

\hline
Conv 1&$56\times 56 \times 64$&$3 \times 3$ & 2\\
\hline
Conv 2&$28\times 28 \times 128$&$3 \times 3$ & 2\\
\hline
Conv 3&$14\times 14 \times 256$&$3 \times 3$ & 2\\
\hline
TransposeConv 1 &$28\times 28 \times 128$&$3 \times 3$ & 2\\
\hline
TransposeConv 2 &$56\times 56 \times 64$&$3 \times 3$ & 2\\
\hline
TransposeConv 3 &$112\times 112 \times 1$&$3 \times 3$ & 2\\
\hline
\end{tabular}}
\end{center}
\end{table}

\begin{table}[h]
\begin{center}
\caption{Architecture of EPN}
\label{tab:EPN}
\setlength{\tabcolsep}{1mm}
{
\begin{tabular}{c|c|c|c}
\hline
\bf{Layer Name}&\bf{Output Size}&\bf{Kernel Size}&\bf{Stride}\\
\hline
Conv 1 & $112\times 112 \times 64$ & $3 \times 3$ & 1\\
\hline
Block 2&$56\times 56 \times 64$&$\left\{\begin{matrix} 3\times 3\\3\times 3 \end{matrix} \right\} \times 2$ & 2\\
\hline
Block 3&$28\times 28 \times 128$&$\left\{\begin{matrix} 3\times 3\\3\times 3 \end{matrix} \right\} \times 2$ & 2\\
\hline
Block 4&$14\times 14 \times 256$&$\left\{\begin{matrix} 3\times 3\\3\times 3 \end{matrix} \right\} \times 2$ & 2\\
\hline
Block 5&$7\times 7 \times 512$&$\left\{\begin{matrix} 3\times 3\\3\times 3 \end{matrix} \right\} \times 2$ & 2\\
\hline
\end{tabular}}
\end{center}
\end{table}

\begin{table}[h]
\begin{center}
\caption{Architecture of the fully connected layers}
\label{tab:FC}
\setlength{\tabcolsep}{1mm}
{
\begin{tabular}{c|c|c|c}
\hline
\bf{Layer Name}&\bf{Input Size}&\bf{Output Size}&\bf{Activation Function}\\
\hline
FC 1& 25088 & 512 & PReLU\\
\hline
FC 2& 512 & 2500 & Sigmoid\\
\hline
\end{tabular}}
\end{center}
\end{table}

\begin{figure*}[t]
\begin{center}
\setlength{\abovecaptionskip}{0pt}
\setlength{\belowcaptionskip}{0pt}
\includegraphics[width=0.8\linewidth]{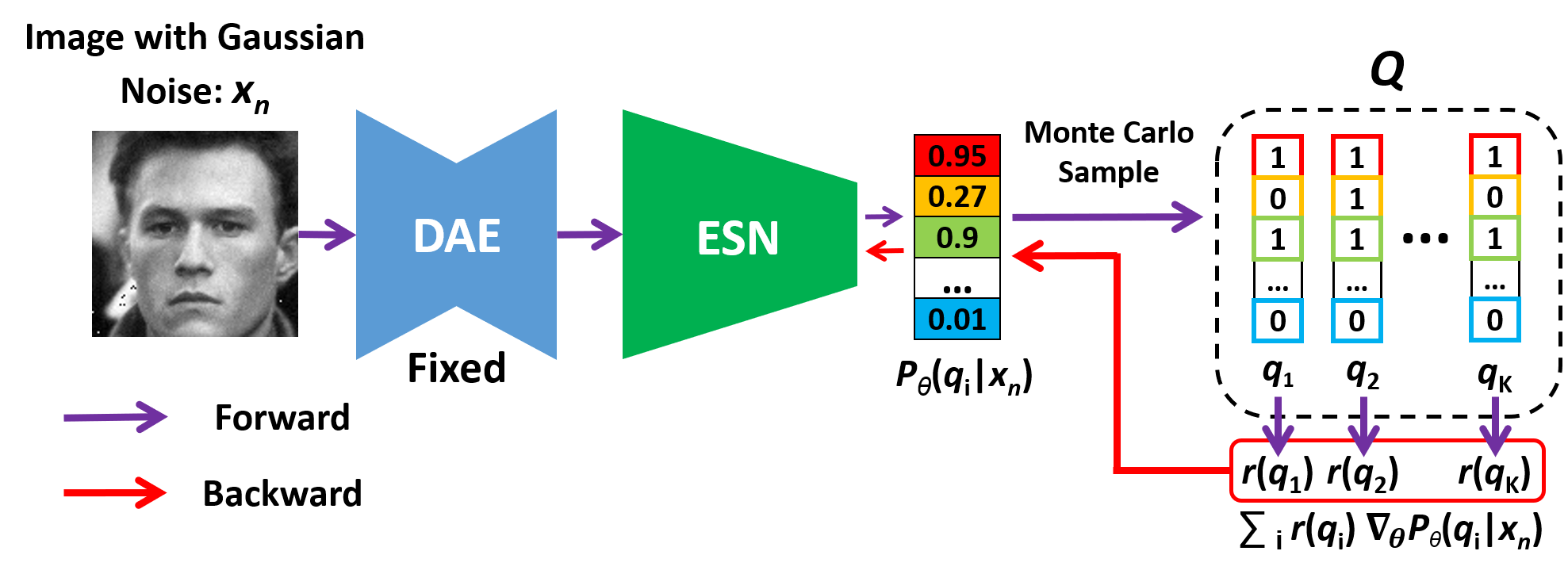}
\end{center}
\setlength{\abovecaptionskip}{0pt}
\setlength{\belowcaptionskip}{0pt}
   \caption{Training of the proposed framework. The pre-trained denoising autoencoder is used to accelerate the convergence of the training process and is frozen during the training. Inspired by the deep reinforcement learning algorithm, policy gradients are adopted to train the model.}
\label{fig:TrainFramework}
\end{figure*}

In the proposed learnable PCA, eigenvector selection is non-differential because of the sampling operation. 
Hence, the framework cannot be trained directly by gradient backpropagation.
Inspired by the deep reinforcement learning algorithm~\cite{Williams1992}\cite{Xin2018Relaxation}, we train the networks by policy gradients, as shown in Fig.~\ref{fig:TrainFramework}.
%

First of all, we define the training object.
For a selection vector $q$ yielded by the framework, it can be regarded as an action, and we can define a reward function to evaluate the action $q$:
\begin{equation}
 r(q) = - ||B_q B_q^T (x_{n}-x_{m}) - (x - x_{m})||_2 - \lambda ||q||_0
\end{equation}
To improve the generalization ability, the input images are contaminated by Gaussian noises during training, where $x_{n}$ is the contaminated input image, $x_m$ is the mean of all clean images in the training set, $x$ is the clean image, and $B_q$ is a matrix consisting of the selected eigenvectors. 
The first term of the reward function evaluates the reconstruction quality.
The second term, which is the zero-norm of $q$, is a regularization term.
The regularization prefers to activate fewer eigenvectors during reconstruction.
Hence, the agent is urged to find a subspace with a lower dimension, and compress the search space of the attack method in \emph{online defense}.
$\lambda$ is a scaling parameter for balancing the reconstruction loss and the regularization.

The goal of training is to maximize the expected reward for the input images:
\begin{equation}
\max_{\theta}~ \mathbb{E}_{q\sim \mathscr{B}_{\theta}(x_{n})}[r(q)]
\end{equation}
where $\mathscr{B}_{\theta}(x_{n})$ is the Bernoulli distributions of $q$. Hence, the goal of training is to minimize the negative expected reward:
\begin{equation}
L(\theta) = -\mathbb{E}_{q\sim \mathscr{B}_{\theta}(x_{n})}[r(q)]
\end{equation}

We can estimate the expected reward by Monte Carlo sampling:
\begin{equation}
\mathbb{E}_{q\sim \mathscr{B}_{\theta}(x_{n})}[r(q)] \approx \mathbb{E}_{q\in Q}[r(q)]
\end{equation}
where $Q$ is a set of $q$ generated by Monte Carlo sampling according to the Bernoulli distributions of $q$:
\begin{equation}
Q = \{q_1, q_2, ..., q_k\} = MC^{P_{\theta}(q|x_{n})}(k)
\end{equation}
Then:
\begin{equation}
\mathbb{E}_{q\sim \mathscr{B}_{\theta}(x_{n})}[r(q)] \approx \mathbb{E}_{q\in Q}[r(q)] = \frac{1}{k}\sum_{i=1}^k r(q_i)P_{\theta}(q_i|x_{n})
\label{equ:estimate}
\end{equation}
where $P_{\theta}(q_i|x_{n})$ is the probability of $q_i$ under the condition of input image $x_{n}$. Hence, the gradient for training can be computed as:
\begin{equation}
\nabla_\theta L(\theta) \approx -\frac{1}{k} \sum_{i=1}^k r(q_i) \nabla_\theta P_{\theta}(q_i|x_{n})
\end{equation}

However, this gradient is not stable during training. To reduce the variance of the gradient, we adopt a baseline strategy~\cite{Williams1992}\cite{Xin2018Relaxation}, which is a widely used method in reinforcement learning. 
For each input image:
\begin{equation}
\nabla_\theta L(\theta) \approx -\frac{1}{k} \sum_{i=1}^k (r(q_i) - r_0) \nabla_\theta P_{\theta}(q_i|x_{n})
\label{equ:gradient}
\end{equation}
where the baseline $r_0$ is independent of the action $q_i$. The baseline $r_0$ does not change the estimation in Eq.~\ref{equ:estimate}, but it can effectively reduce the variance of the gradient. We select the average of the rewards of each minibatch as the baseline. Then, the networks can be trained based on the gradient computed in Eq.~\ref{equ:gradient}.


\subsection{Training Details}
\label{sec:MetDet}
Training details of the proposed framework are provided in this subsection.
The size of the input image is $112\time 112$.
All the input images are preprocessed using the same method as that in ArcFace~\cite{Deng2018ArcFace}.
The input images are transformed into grayscale with only one channel for projection.

Only the first 2500 eigenvectors are considered. The probabilities of other eigenvectors are fixed to 0.
The intensity of perturbation is represented by the ratio of the scale of the perturbation and scale of the clean image:
 \begin{equation}
 I(\eta) = \frac{||\eta||}{||x||}
 \end{equation}
The intensity of the Gaussian noise is set to 0.04 during training. The pretrained DAE is trained on CASIA WebFace~\cite{Yi2014Deep} perturbed by Gaussian noise with the same intensity.

Stochastic gradient descent with momentum is adopted for optimization. The basic learning rate is 0.01, and the learning rate decreases by 50\% every 20 epochs. The momentum is fixed to 0.9. $\lambda$ in the reward function is set to 0.015.
Without elaborate adjusting, these hyper-parameters are selected according to the conventional practice of deep learning model training.



\section{Experiments}
\label{Experiment}

\begin{table*}[t]
\begin{center}
\setlength{\abovecaptionskip}{0pt}
\setlength{\belowcaptionskip}{0pt}
\caption{EER on LFW.  The proposed method performs better in all three attacking cases and provides comparative performance on clean images at the same time.}
\label{tab:ICLFW}
\setlength{\tabcolsep}{6mm}
{
\begin{tabular}{c|c|c|c|c|c}
\hline
\multicolumn{2}{c|}{\bf{Attacking Method}}  & \bf{Clean} & \bf{FGSM} & \bf{DeepFool} & \bf{PGD} \\
\hline
\multirow{9}*{Offline} &No Defense  & 0.44\% & 41.97\% &89.49\% & 99.71\%  \\
\cline{2-6}
& Quilting~\cite{Moosavi2018Divide}  & 8.77\% & 9.04\% &25.10\% & 45.79\%  \\
\cline{2-6}
 & TVM~\cite{Guo2017Countering}  & 2.95\% & 19.94\% &73.21\% & 96.62\%  \\
\cline{2-6}
 & PixelDefend~\cite{Song2018PixelDefend}  & 2.05\% & 18.02\% &70.26\% & 97.58\%  \\
\cline{2-6}
 & MagNet~\cite{Meng2017MagNet} & 1.52\% & 7.84\% &14.56\% &46.07\%  \\
\cline{2-6}
 & Adversarial MTER~\cite{Zhong2019adversarial}  & 2.62\% & 10.03\% &24.89\% & 61.06\%  \\
\cline{2-6}
& HGD~\cite{Liao2018Defense} & 1.08\% & 17.35\% & 20.48\% &49.69\%  \\
\cline{2-6}
& Xie et al.~\cite{xie2019feature} & \bf{0.94}\% & 20.29\% &28.96\% & 31.27\%  \\
\cline{2-6}
 & Ours  & 1.95\% & \bf{6.15}\% &\bf{7.86}\% & \bf{29.77}\%  \\

\hline
\hline

\multirow{6}*{Online} &No Defense  & 0.44\% & 41.97\% &89.49\% & 99.71\%  \\
\cline{2-6}
 & MagNet~\cite{Meng2017MagNet} & 1.52\% & 41.64\% &51.88\% &98.74\%  \\
\cline{2-6}
 & Adversarial MTER~\cite{Zhong2019adversarial}  & 2.62\% & 69.21\% &73.59\% & 98.08\%  \\
\cline{2-6}
& HGD~\cite{Liao2018Defense} & 1.08\% & 36.67\% & 80.75\% &89.64\%  \\
\cline{2-6}
& Xie et al.~\cite{xie2019feature} & \bf{0.94}\% & 37.99\% &82.27\% & 87.66\%  \\
\cline{2-6}
 & Ours  & 1.95\% & \bf{14.51}\% &\bf{34.66}\% & \bf{63.32}\%  \\

\hline
\end{tabular}}
\end{center}
\end{table*}

\begin{table*}[t]
\begin{center}
\setlength{\abovecaptionskip}{0pt}
\setlength{\belowcaptionskip}{0pt}
\caption{EER on MegaFace. The average EERs are reported. The $\pm$ shows 95\% confidence intervals.}
\label{tab:ICMega}
\setlength{\tabcolsep}{3mm}
{
\begin{tabular}{c|c|c|c|c|c}
\hline
\multicolumn{2}{c|}{\bf{Attacking Method}}              & \bf{Clean} & \bf{FGSM} & \bf{DeepFool} & \bf{PGD} \\
\hline
\multirow{9}*{Offline} &No Defense                          & 1.31\%                          &  50.87\%  &  95.41\%    &  99.09\%   \\
\cline{2-6}
& Quilting~\cite{Moosavi2018Divide}                         & 14.23\%                       & 18.08\%     &  20.54\%   & 47.13\%  \\
\cline{2-6}
 & TVM~\cite{Guo2017Countering}                            & 3.66\%                          & 21.59\%     &. 76.24\%    & 95.12\%  \\
\cline{2-6}
 & PixelDefend~\cite{Song2018PixelDefend}             & 2.43\%                            & 25.04\%     &  77.62\%    & 94.20\%  \\
\cline{2-6}
 & MagNet~\cite{Meng2017MagNet}                          & 2.68\%                    & 10.56\%       & 23.95\%     &  47.07\%  \\
\cline{2-6}
 & Adversarial MTER~\cite{Zhong2019adversarial}  & 3.08\%                           & 11.50\%       &  33.44\%     & 66.95\%  \\
\cline{2-6}
& HGD~\cite{Liao2018Defense}                                & 2.44\%                            & 16.15\%       & 29.09\%     &  52.27\%  \\
\cline{2-6}
& Xie et al.~\cite{xie2019feature}                                & \bf{1.93}\%                     & 15.96\%       &  34.57\%     & 49.09\%  \\
\cline{2-6}
 & Ours                                                                           & 3.78\%$\pm$1.01\%    & \bf{7.82}\%$\pm$0.53\%     &    \bf{9.20}\%$\pm$0.96\%    &    \bf{33.37}\%$\pm$1.97\%  \\

\hline
\hline

\multirow{6}*{Online} &No Defense                           & 1.31\%                          &  50.87\%  &  95.41\%    &  99.10\%   \\
\cline{2-6}
 & MagNet~\cite{Meng2017MagNet}                          & 2.68\%                  & 63.69\%   &  76.23\%      &99.09\%  \\
\cline{2-6}
 & Adversarial MTER~\cite{Zhong2019adversarial}  & 3.08\%                          &   60.45\%   &78.24\% & 97.91\%  \\
\cline{2-6}
& HGD~\cite{Liao2018Defense}                                & 2.44\%                            & 76.29\%       & 79.04\%     &  96.18\%  \\
\cline{2-6}
& Xie et al.~\cite{xie2019feature}                                & \bf{1.93}\%                     & 60.57\%       &  74.59\%     & 94.33\%  \\
\cline{2-6}
 & Ours                                                                          & 3.78\%$\pm$1.01\%     & \bf{22.29}\%$\pm$1.86\%  &\bf{42.53}\%$\pm$1.62\% & \bf{61.12}$\pm$3.11\%  \\

\hline
\end{tabular}}
\end{center}
\end{table*}

\subsection{Defense Protocols}

Given the characteristics of face recognition, there are two kinds of adversarial defense protocols:

\emph{1) Offline defense}: The adversarial examples are generated by attackers before the defense strategy is deployed. This means that the attackers are not aware of any defense strategy and generate adversarial examples based on the unsecured recognition model. Defense methods are evaluated on these adversarial examples. Formally, the adversarial perturbation of offline defense is generated by the following optimization:
\begin{equation}
\epsilon_{off} = \mathop{\arg\max}\limits_{\epsilon} \mathscr{L}_{M}(x+\epsilon)
\end{equation}
where $\epsilon$ is the perturbation, $x$ is the clean image, $\mathscr{L}_{M}$ is the optimization target based on recognition model $M$, and $M$ is the mode without any defense.

\emph{2) Online defense}: In this scenario, the attackers are aware of the in-service defenders so that the attackers can attempt to dupe the defenders. This means that either the attacker successfully fools the defended recognition model or the defender wins. Formally, the adversarial perturbation of online defense is generated by the following optimization:
\begin{equation}
\epsilon_{on} = \mathop{\arg\max}\limits_{\epsilon} \mathscr{L}_{M^*}(x+\epsilon)
\end{equation}
where $M^*$ is a recognition model with defense.
For the proposed method, we adopt the reparameterization technology, which is proposed in VAE~\cite{kingma2013auto}, to generate the gradient for the sampling operation.
It has been proved to be an effective and efficient way to generate the gradient for sampling operations.

For defense methods, online defense is obviously more difficult than offline defense. More importantly, online defense is more common in practice because $M^*$, rather than $M$, is the target of the attackers once the defense method has been deployed. Hence, the performance of online defense is more important for evaluation.


\begin{figure}[ht]
\begin{center}
\setlength{\abovecaptionskip}{0pt}
\setlength{\belowcaptionskip}{0pt}
\includegraphics[width=\linewidth]{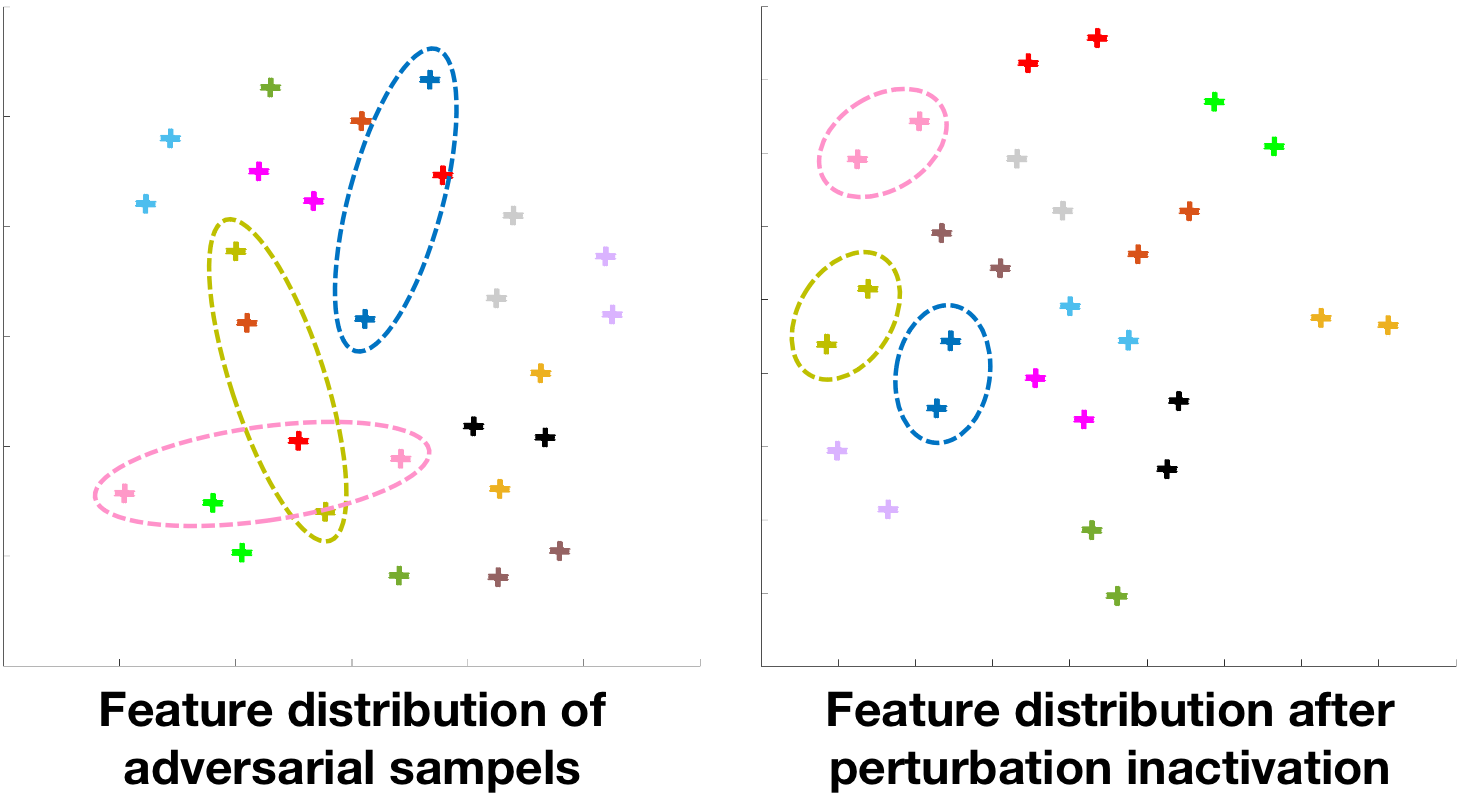}
\end{center}
\setlength{\abovecaptionskip}{0pt}
\setlength{\belowcaptionskip}{0pt}
   \caption{Feature distribution of adversarial samples produced by FGSM before and after applying the perturbation inactivation (visualized by tSNE). The same color denotes the positive pairs. The positive pairs are drawn closer by the perturbation inactivation. Beat viewed in color.}
\label{fig:tsne}
\end{figure}

\subsection{Evaluation under General Attacking Methods}
\label{ExpIC}
We select three attacking methods meant for the general classification in this subsection to evaluate the proposed framework: FGSM~\cite{Goodfellow2014Explaining}, which is a popular attacking method, DeepFool~\cite{Moosavi2016DeepFool} and PGD~\cite{Madry2017Towards}, which is a powerful approach proposed recently.
ArcFace~\cite{Deng2018ArcFace} is used as the recognition model.
The distances in feature space between the facial image pairs are used as their objectives. For positive pairs, the goal of the attacking methods is to enlarge the distances, while for negative pairs, the goal is to reduce the distances.

The experiments are launched on two datasets: Labeled Faces in the Wild (LFW)~\cite{Learned2016Labeled} and MegaFace~\cite{Kemelmacher2016The}. 
For LFW, we follow the official benchmark protocol\footnote{http://vis-www.cs.umass.edu/lfw/pairs.txt}, where 3,000 positive pairs and 3,000 negative pairs of images are selected for face verification.
For MegaFace, 80 IDs in the subset \emph{facescrub} with more than 50 images per subject are selected, and 10 images of each identity are randomly selected for testing.
Hence, there are 7,200 positive pairs and 632,000 negative pairs. 
To provide more statistically significant performance, we randomly select the images from the 80 identities five times and the statistical results are reported.
Adversarial perturbations are added to one image of the pairs and the other image is fixed during testing. 
We randomly select additional 5000 images of each dataset for eigenvector generation.
There is no overlap between these 5000 images and the test pairs.
The proposed framework is trained on CASIA-WebFace~\cite{Yi2014Deep}.

Seven methods of defense are selected for comparison in this subsection, including Quilting~\cite{Moosavi2018Divide}, TVM~\cite{Guo2017Countering}, PixelDefend~\cite{Song2018PixelDefend}, and MagNet~\cite{Meng2017MagNet}.
Adversarial MTER~\cite{Zhong2019adversarial}, which is an adversarial training-based method for face recognition as mentioned in Section~\ref{RelateWork}, is compared in this subsection. 
Two denoising-based methods, HGD~\cite{Liao2018Defense}, and Xie. et al~\cite{xie2019feature}, are also taken into consideration.
The official implementation of adversarial MTER and HGD is used in our experiments.
PixelDefend, MagNet, and Xie. et al are trained from scratch on the same dataset with the proposed framework for a fair comparison.


We adopt EER (equal error rate) of face verification as the index of performance in this subsection. The EER is computed when the intensity of adversarial noises is 0.04.
The computational complexity of the first three defense methods is too high to be applied under an online defense protocol. They are thus evaluated under an offline protocol. The other methods are evaluated under both offline and online protocols.

\begin{figure*}[ht]
\begin{center}
\setlength{\abovecaptionskip}{0pt}
\setlength{\belowcaptionskip}{0pt}
\includegraphics[width=\linewidth]{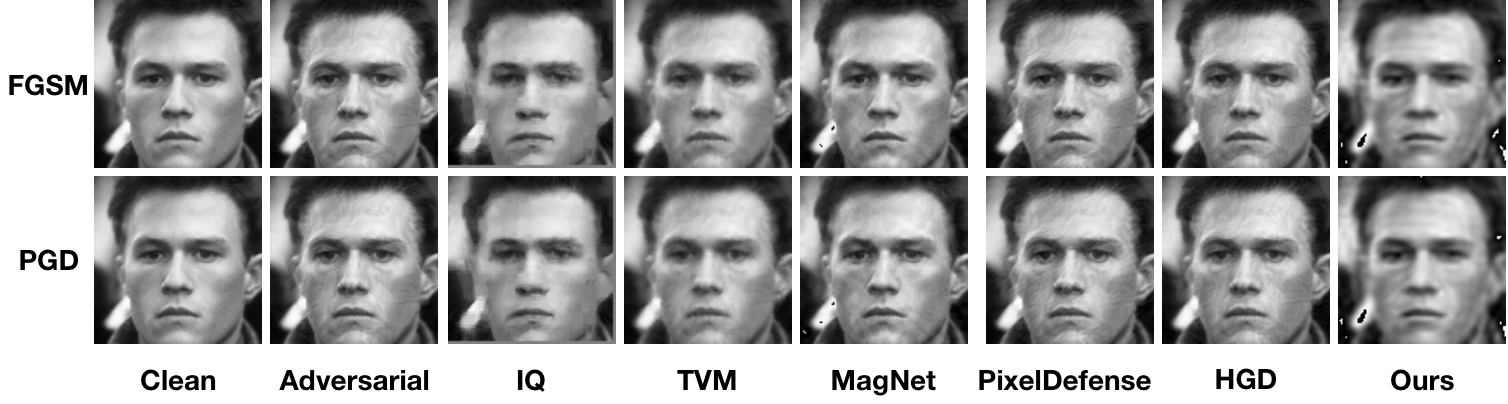}
\end{center}
\setlength{\abovecaptionskip}{0pt}
\setlength{\belowcaptionskip}{0pt}
   \caption{Examples of reconstructed images. The adversarial perturbations are removed by our method while the primary information of the original face images is recovered.}
\label{fig:Recon}
\end{figure*}

The results on LFW are shown in Tab.~\ref{tab:ICLFW}. The results on MegaFace are shown in Tab.~\ref{tab:ICMega}.
For the results on MegaFace, the average EERs are reported.
The $\pm$ shows 95\% confidence intervals.
The feature distributions of the adversarial samples before and after the proposed perturbation inactivation are shown in Fig~\ref{fig:tsne}.
The samples of the same subject (positive pairs) are drawn closer by the perturbation inactivation.
The examples of reconstructed images are shown in Fig.~\ref{fig:Recon}.
The adversarial perturbations are removed by our method while the primary information of the original face images is recovered.
The quantitative results show that the proposed method performs better in all three attacking cases and provides comparative performance on clean images at the same time.
More importantly, the gaps between the offline and online defense of the proposed method are smaller than in other methods. 
The adversarial perturbations are inactivated by projection into the estimation of the immune space where the recognition model is robust to perturbations. 
Hence, it is difficult for attackers to find effective perturbations in both offline and online defense protocols.
%


\subsection{Evaluation under Attacking Methods for Face Recognition}
\label{ExpFR}
Two kinds of attacking methods tailored for face recognition are adopted to evaluate the proposed framework in this subsection. ArcFace~\cite{Deng2018ArcFace} is also used as the recognition model. 


\subsubsection{Black-box Attack for Face Recognition}
\label{ExpFRBlack}

The first attacking method is a black-box attacking method of evolutionary attack~\cite{Dong2019Efficient}. This attacking method iteratively reduces the difference between the adversarial image and clean image on the basis of successfully attacking. 
Hence, we report the average distortion between the adversarial images and clean images as the index of performance. At the same step of the iteration, a higher average distortion indicates better performance of the defense method.

Only online defense protocols are adopted in this experiment. 
Two kinds of attacks are taken into consideration: dodging and impersonation.
Dodging means that the goal of attacking is to reduce the similarity of positive pairs to confuse the recognition model.
By contrast, impersonation means that the goal of attacking is to increase the similarity of negative pairs.
The threshold of similarity is set to 0.2 (the possible range is $[-1, 1]$), which means that image pairs with a similarity higher than 0.2 are predicted as positive pairs.
LFW and MegaFace are used in this experiment.
The selection of test pairs is the same with Section~\ref{ExpIC} (if the recognition model wrongly predicts a clean pair, this pair will be removed to apply the black-box attack).

The results on LFW are shown in Tab.~\ref{tab:FRBlackLFW}. The results on MegaFace are shown in Tab.~\ref{tab:FRBlackMega}.
The average results are reported on MegaFace.
The results demonstrate the advantages of the proposed framework. 
The average residuals of the proposed framework on both two datasets are larger than those obtained by other methods, and the residuals decrease much more slowly than those of the other methods as the number of the iteration steps increases.

\begin{table}[t]
\begin{center}
\setlength{\abovecaptionskip}{0pt}
\setlength{\belowcaptionskip}{0pt}
\caption{Average distortion (MSE) on LFW under the black-box attack. At the same step of iteration, a higher average distortion indicates better performance of the defense method.}
\label{tab:FRBlackLFW}
{
\begin{tabular}{c|c|c|c|c}
\hline
\multicolumn{2}{c|}{\bf{Number of Iteration Steps}}  & \bf{1000} & \bf{5000} & \bf{10000} \\
\hline
\multirow{6}*{Dodging} & No Defense                       & 3.1e-3 & 2.5e-4 & 8.9e-5  \\
\cline{2-5}
&MagNet~\cite{Meng2017MagNet}                            &  1.0e-2 & 7.7e-3 & 5.8e-3  \\
\cline{2-5}
&Adversarial MTER~\cite{Zhong2019adversarial}    & 9.9e-3 & 6.5e-3 & 4.9e-3  \\
\cline{2-5}
& HGD~\cite{Liao2018Defense}                                & 8.7e-3 & 6.3e-3 & 3.9e-3\\
\cline{2-5}
& Xie et al.~\cite{xie2019feature}                                & 7.4e-3 & 5.8e-3 & 2.7e-3 \\
\cline{2-5}
&Ours                                                                             & \bf{7.5e-2} & \bf{5.5e-2} & \bf{3.1e-2}  \\
\hline
\multirow{6}*{Impersonation} & No Defense             &  2.4e-3 & 2.4e-4 & 5.9e-5  \\
\cline{2-5}
&MagNet~\cite{Meng2017MagNet}                            & 8.3e-3 & 5.7e-3 & 3.1e-3  \\
\cline{2-5}
&Adversarial MTER~\cite{Zhong2019adversarial}    & 1.7e-3 & 9.3e-4 & 6.1e-4  \\
\cline{2-5}
& HGD~\cite{Liao2018Defense}                                & 1.1e-3 & 7.2e-4 & 4.8e-4 \\
\cline{2-5}
& Xie et al.~\cite{xie2019feature}                                & 9.7e-4 & 6.6e-4 & 4.1e-4 \\
\cline{2-5}
&Ours                                                                             &  \bf{4.6e-2} & \bf{3.6e-2} & \bf{2.9e-2}  \\
\hline
\end{tabular}}
\end{center}
\end{table}

\begin{table}[t]
\begin{center}
\setlength{\abovecaptionskip}{0pt}
\setlength{\belowcaptionskip}{0pt}
\caption{Average distortion (MSE) on MegaFace. The results on average are reported.}
\label{tab:FRBlackMega}
{
\begin{tabular}{c|c|c|c|c}
\hline
\multicolumn{2}{c|}{\bf{Number of Iteration Steps}}  & \bf{1000} & \bf{5000} & \bf{10000} \\
\hline
\multirow{6}*{Dodging} & No Defense                    & 3.5e-3 &  8.5e-4&   9.7e-5\\
\cline{2-5}
&MagNet                                                                     & 9.8e-2 & 6.5e-3 & 2.2e-3  \\
\cline{2-5}
&Adversarial MTER~\cite{Zhong2019adversarial} & 9.7e-3 & 6.6e-3 & 5.1e-3  \\
\cline{2-5}
& HGD~\cite{Liao2018Defense}                                & 8.8e-2 & 6.0e-3 & 1.9e-3 \\
\cline{2-5}
& Xie et al.~\cite{xie2019feature}                                & 7.2e-3 & 5.5e-3 & 2.3e-3 \\
\cline{2-5}
&Ours                                                                          & \bf{9.2e-2} & \bf{8.4e-2} & \bf{6.7e-2}  \\
\hline

\multirow{6}*{Impersonation} & No Defense           & 2.4e-3 &  1.1e-4& 5.5e-5  \\
\cline{2-5}
&MagNet                                                                     & 7.9e-3 & 2.1e-3 & 1.4e-3\\
\cline{2-5}
&Adversarial MTER~\cite{Zhong2019adversarial} & 7.9e-3 & 4.6e-3 & 2.9e-3  \\
\cline{2-5}
& HGD~\cite{Liao2018Defense}                                & 7.2e-3 & 1.7e-3 & 8.4e-4 \\
\cline{2-5}
& Xie et al.~\cite{xie2019feature}                                & 6.5e-3 & 3.7e-3 & 1.8e-3 \\
\cline{2-5}
&Ours                                                                           & \bf{7.7e-2} & \bf{6.8e-2} & \bf{6.2e-2}  \\
\hline
\end{tabular}}
\end{center}
\end{table}


\subsubsection{Sticker Attack for Face Recognition}
\label{ExpFRSticker}

Another attacking method tailored for face recognition is the sticker attack~\cite{Komkov2013AdvHat}. In this kind of attack, a predefined part of the face is covered by an adversarial sticker to confuse the face recognition model. For example, different people with the same sticker are recognized as the same person. Different from the other attacking methods, the sticker attack does not limit the scale of the perturbation inside the predefined part, which means that it is harder to defend against.
\begin{figure}[h]
\begin{center}
\setlength{\abovecaptionskip}{0pt}
\setlength{\belowcaptionskip}{0pt}
\includegraphics[width=0.8\linewidth]{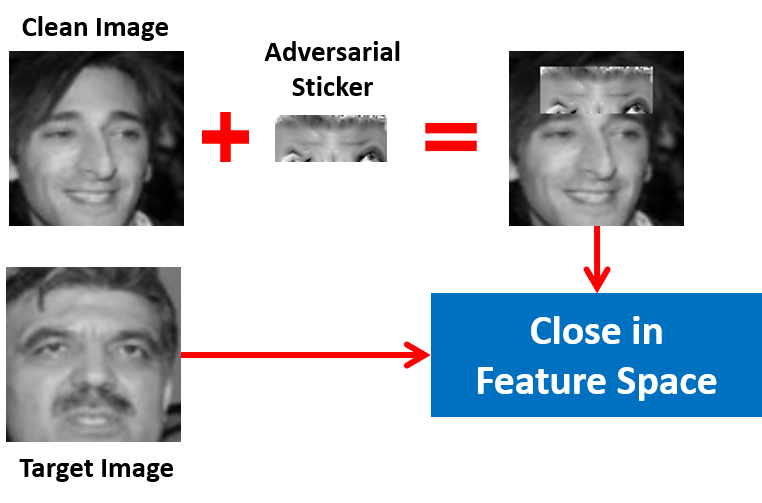}
\end{center}
\setlength{\abovecaptionskip}{0pt}
\setlength{\belowcaptionskip}{0pt}
   \caption{Adversarial sticker attack. The special region of the face image is covered by the adversarial sticker to fool the recognition model.}
\label{fig:sticker}
\end{figure}

In this experiment, we trained an adversarial sticker in the same manner as~\cite{Komkov2013AdvHat} to cover the forehead of faces. The goal of the attack is to make the 1000 face images of 1000 different persons tend to be recognized as the target person as shown in Fig.~\ref{fig:sticker}. The 1000 images and the target image are selected from LFW. The size of the sticker is $20\times 72$ pixels. The matching threshold of similarity is set to 0.2 (the possible range is $[-1, 1]$) so that most positive pairs can be recognized correctly by the recognition model (ArcFace) while the false negative rate is acceptable (TPR=99.89\%, FNR=2.09\%) on clean images.
It is a strict setting for defense methods.
The similarities between the 1000 face images and the target image should be smaller than the threshold because they are obtained from different persons. If the similarity between a face image and the target image is smaller than the threshold, the result is considered to be correct. By contrast, if the similarity is larger than the threshold, the result is wrong. The error rate is adopted to evaluate the performance in this experiment. All clean images of 1000 different persons can be recognized correctly. Both stickers of offline defense and online defense are used to evaluate the defense methods.

\begin{table}[t]
\begin{center}
\setlength{\abovecaptionskip}{0pt}
\setlength{\belowcaptionskip}{0pt}
\caption{Error rate under adversarial sticker attack. The proposed framework is better than other methods. The gap between offline and online defense is also the smallest.}
\label{tab:sticker}
\setlength{\tabcolsep}{5mm}
{
\begin{tabular}{c|c|c}
\hline
\bf{Defense Method} & \bf{Offline} & \bf{Online} \\
\hline
No Defense & 74.4\% & 74.4\%   \\
\hline
Quilting~\cite{Moosavi2018Divide} & 7.8\% & - \\
\hline
TVM~\cite{Guo2017Countering} & 14.0\% & -  \\
\hline
PixelDefend~\cite{Song2018PixelDefend} &7.3\% & -   \\
\hline
MagNet~\cite{Meng2017MagNet} & 4.1\% & 34.9\%   \\
\hline
Adversarial MTER~\cite{Zhong2019adversarial} & 6.6\% & 48.5\%   \\
\hline
HGD~\cite{Liao2018Defense}   & 10.9\% & 45.7\% \\
\hline
Xie et al.~\cite{xie2019feature}     & 9.3\% & 54.2\% \\
\hline 
Ours & \bf{2.4\%} & \bf{7.2}\% \\
\hline
\end{tabular}}
\end{center}
\end{table}

The results are shown in Tabel~\ref{tab:sticker}. It is observed from the results that the proposed framework is better than other methods. The gap between offline and online defense is 4.8\%, while those of the compared methods are much larger.
Not surprisingly, the performance of TVM is the worst in offline defense, since the TVM loss has been taken into consideration during the training of the adversarial sticker. Hence, the effectiveness of the defense method based on TVM is quite limited.

The adversarial stickers are inactivated by projection into the immune space.
Thus, the adverse impact of the adversarial sticker is significantly mitigated, while the principal information of the original image is maintained.
%


\subsection{Evaluation of Commercial APIs}

\begin{table*}[t]
\begin{center}
\setlength{\abovecaptionskip}{0pt}
\setlength{\belowcaptionskip}{0pt}
\caption{Recognition performance on TALFW~\cite{TALFW}. Our defense method leads to significant improvements for all of the commercial APIs and the ArcFace. The adversarial training strategy only achieves negligible improvement compared to the unsecured ArcFace.}
\label{tab:talfw}
\setlength{\tabcolsep}{2mm}
{
\begin{tabular}{c|c|c|c|c|c}
\hline
 & \bf{AUC$\times$100\%} & \bf{TAR@FAR=10\%} & \bf{TAR@FAR=1\%}  & \bf{TAR@FAR=0.1\%} & \bf{100\% - EER}  \\
\hline
Amazont~\cite{AmazonAPI} & 73.36\% & 42.5\% & 10.17\% & 2.9\%   & 68.13\% \\
Microsoft~\cite{MicrosoftAPI} & 74.77\% & 40.53\% & 12.97\% & 5.4\%   & 68.33\% \\
Baidu~\cite{BaiduAPI}       & 74.45\% & 37.7\%   & 11.2\%   & 2.5\%   & 68\%      \\
Face++~\cite{FaceppAPI}     & 78.68\% & 41.73\% & 11.57\% & 3.13\% & 71.73\%  \\
ArcFace~\cite{Deng2018ArcFace}    & 60.48\% & 21.77\% & 3.4\%     & 0.53\% & 57.47\%  \\

\hline
Adversarial Training~\cite{Goodfellow2014Explaining}  & 61.29\% & 21.67\% & 5.23\% & 1.47\%   & 57.8\% \\

Ours + Amazon~\cite{AmazonAPI}  & \bf{98.48\%}       & \bf{96.53\%}         & \bf{85.83\%} & \bf{69.02\%}           & \bf{94.23\%} \\
Ours + Microsoft~\cite{MicrosoftAPI}  & 96.18\%       & 89.83\%         & 65.44\% & 40.28\%           & 89.96\% \\
Ours + Baidu~\cite{BaiduAPI}        & 96.02\%       & 88.34\%         & 64.88\%        & 40.96\%           & 89.13\%         \\
Ours + Face++~\cite{FaceppAPI}      & 96.4\%  & 89.86\% & 63.81\%         & 43.72\%   & 89.89\%         \\

Ours + ArcFace~\cite{Deng2018ArcFace}     & 94.63\%   & 86.3\% & 57.4\%  & 28.47\%   & 87.9\% \\
\hline
\end{tabular}}
\end{center}
\end{table*}

\begin{figure}[t]
\begin{center}
\setlength{\abovecaptionskip}{0pt}
\setlength{\belowcaptionskip}{0pt}
\includegraphics[width=\linewidth]{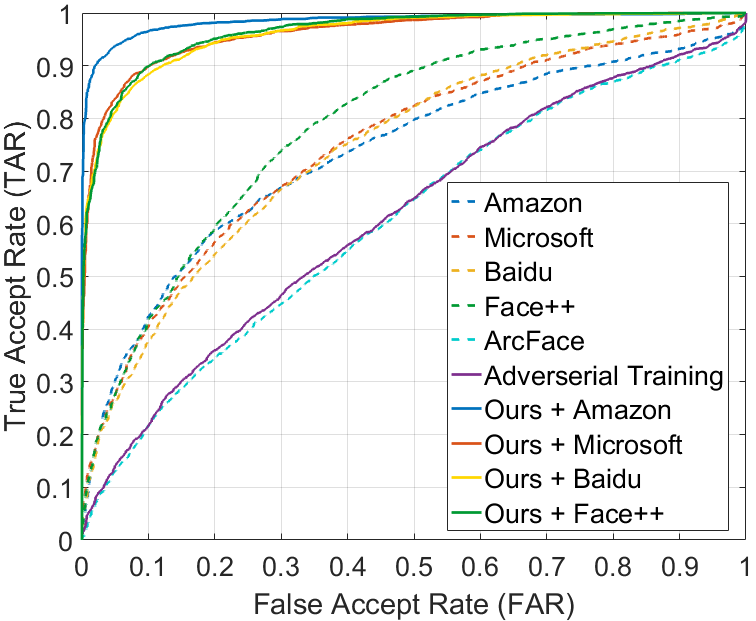}
\end{center}
\setlength{\abovecaptionskip}{0pt}
\setlength{\belowcaptionskip}{0pt}
   \caption{ROC curves on TALFW~\cite{TALFW}. Our defense method leads to significant improvements for all of the commercial APIs and the ArcFace. The strong adversarial perturbations in the facial images are inactivated effectively by our method. The adversarial training strategy only achieves negligible improvement compared to the unsecured ArcFace. Best viewed in color.}
\label{fig:talfw}
\end{figure}

The proposed method defends against adversarial attack by inactivating the adversarial perturbations before the recognition model.
Hence, it can be applied to the existing face recognition system directly.
To verify its interoperability, the proposed method is applied to four commercial face recognition APIs, namely, Amazon~\cite{AmazonAPI}, Microsoft~\cite{MicrosoftAPI}, Baidu~\cite{BaiduAPI}, and Face++~\cite{FaceppAPI}, to test its effects.

TALFW~\cite{TALFW}, which is an adversarial facial image database, is used as the testing database in this experiment.
TALFW is created based on the well-known face recognition benchmark LFW by adding adversarial perturbations to the original LFW images.
The adversarial perturbations are generated by DFANet~\cite{Zhong2020Towards}, which is a strong adversarial attack method designed for face recognition.
Since the only difference is the adversarial perturbations, the evaluation protocol of TALFW is exactly the same as that of LFW.
The adversarial facial image pairs in TALFW can successfully attack all four commercial APIs.

The testing facial images in TALFW are projected into the estimated subspaces by the proposed method to inactivate the adversarial perturbations. Then, they are fed to the APIs for evaluation.
Our defense method is applied in this experiment directly without any extra training.
ArcFace (ResNet-50)~\cite{Deng2018ArcFace}, which is a state-of-the-art face recognition model, is tested on TALFW for comparison.
Moreover, as a common strategy for adversarial defense, adversarial training~\cite{Goodfellow2014Explaining} is applied in this experiment.
The feature extractor is trained by the adversarial samples generated by IFTGSM~\cite{Zhong2019adversarial}, which is an attack method based on iterative FGSM~\cite{Goodfellow2014Explaining}.

The results are shown in Fig.~\ref{fig:talfw} and Tab.~\ref{tab:talfw}.
Our defense method shows significant improvements compared to all of the commercial APIs and ArcFace.
The EERs of the commercial APIs are decreased by an average of 21.76\%, and the EER of ArcFace is decreased by 30.43\%.
It is verified that the strong adversarial perturbations in the facial images are inactivated effectively by our method.
The results demonstrate the superior generalization capacity and interoperability of our method with different recognition models.

On the other hand, the adversarial training strategy only achieves negligible improvement compared to the recognition model without any defense.
The results show that the generalization capacity of the adversarial training strategy to unseen adversarial attacking is quite limited.

\begin{figure}[t]
\begin{center}
\setlength{\abovecaptionskip}{0pt}
\setlength{\belowcaptionskip}{0pt}
\includegraphics[width=0.7\linewidth]{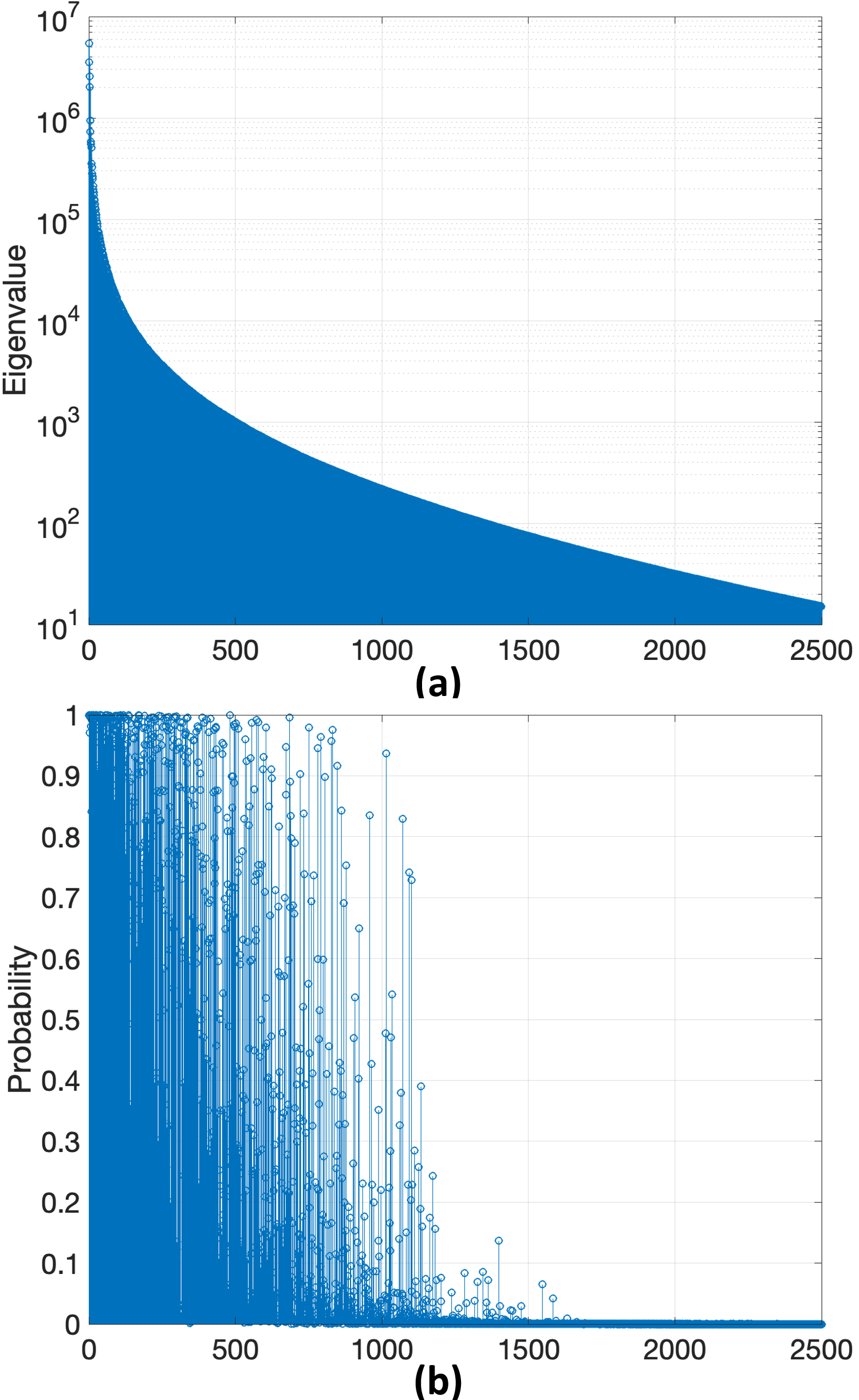}
\end{center}
\setlength{\abovecaptionskip}{0pt}
\setlength{\belowcaptionskip}{0pt}
   \caption{\textbf{(a)} Eigenvalues of LFW. Only the 2500 maximal eigenvalues are shown \textbf{(b)} Selected probabilities of these eigenvectors yielded by the agent for an image from LFW. The eigenvectors with low eigenvalues may have high probabilities to be selected for a special sample and vice versa.}
\label{fig:ValProb}
\end{figure}


\subsection{Cross-domain Evaluation}
\label{CrossDomain}

The generalization ability on different domains of the proposed method is evaluated in this subsection.
It is vital in cross-domain applications since the eigenvectors are necessary during inference and the eigenvectors are generated offline.
Firstly, the proposed method with eigenvectors generated on LFW is evaluated on MegaFace.
The selected pairs are the same as Section~\ref{ExpIC}.
Then, to explore how sensitive the proposed method is to variations in facial poses, the model with eigenvectors generated on LFW is evaluated on Celebrities in Frontal-Profile in the Wild (CFP-FP)~\cite{CFP-Dataset}, which is a collection of face images of celebrities in frontal and profile views.
We follow the official evaluation benchmark of CFP-FP, which contains 7,000 pairs.

The results are shown in Tab.~\ref{tab:CorssDomain}.
The results show that the performance of cross-domain evaluation declines compared to the performance in Tab.~\ref{tab:ICMega}.
However, the results are still better than the comparing methods in Tab.~\ref{tab:ICMega}.
It is demonstrated that the generalization ability on different domains of the proposed method is acceptable.
Actually, the classical PCA is not robust to the domain gap (e.g. camera type, lighting).
However, in the proposed learnable PCA, the eigenvector selection network (ESN) provides the generalization ability across different domains.

Meanwhile, there is a significant gap between the performance of the proposed method and the non-defended recognition model on the clean images of CFP-FP.
This phenomenon indicates that the face images with large facial poses are challenging for the proposed method if the eigenvectors are generated on frontal images.

\begin{table*}[t]
\begin{center}
\setlength{\abovecaptionskip}{0pt}
\setlength{\belowcaptionskip}{0pt}
\caption{Cross-domain evaluation. The average EERs are reported on MegaFace. The $\pm$ shows 95\% confidence intervals. It is demonstrated that the generalization ability on different domains of the proposed method is acceptable. Even though the classical PCA is not robust to the domain gap, the eigenvector selection network (ESN) provides the generalization ability across different domains.}
\label{tab:CorssDomain}
\setlength{\tabcolsep}{4mm}
{
\begin{tabular}{c|c|c|c|c|c|c}
\hline
\multirow{2}*{} &\multicolumn{3}{c|}{\bf{MegaFace}}              & \multicolumn{3}{c}{\bf{CFP-FP}} \\
\cline{2-7}
                       & \bf{Clean}  & \bf{FGSM} & \bf{PGD}  & \bf{Clean}  & \bf{FGSM} & \bf{PGD}\\
\hline
No Defense   & 1.31\%        &  50.87\%      &  99.09\%     & 3.29\%       & 92.01\%       & 99.33\%\\
\hline
Ours-Offline & 10.41\%$\pm$2.48\%     &    17.21\%$\pm$2.01\%    &    36.63\%$\pm$2.42\%   & 20.28\% & 29.14\%   & 42.69\% \\
\hline
Ours-Online & 10.41\%$\pm$2.48\%     &    44.72\%$\pm$2.65\%    &    83.57\%$\pm$0.85\%   & 20.28\%  & 59.77\% & 79.25\% \\
\hline
\end{tabular}}
\end{center}
\end{table*}


\subsection{Differences from Obfuscated Gradient Methods}
\label{ObfGrad}

Obfuscated gradient methods are a kind of defense strategy that leverages gradient masking so that it does not have useful gradients for generating adversarial examples.
There are three ways to realize this defense strategy: shattered gradients, stochastic gradients, and exploding \& vanishing gradients~\cite{athalye2018obfuscated}.
However, this kind of defense strategy is not reliable.
It is proven that the obfuscated gradient methods can be circumvented~\cite{athalye2018obfuscated}.

There is a sampling operation in the proposed method, which may cause misunderstanding that it is similar to the obfuscated gradient method (stochastic gradient).
In this subsection, we discuss the differences between the proposed method and the obfuscated gradient methods.
There are two key differences between them:

\begin{itemize}

\item Their fundamentals are different. 
The stochastic gradient methods rely on the randomness of the network itself or the input samples.
The gradient direction would be disturbed during adversarial attacking.
On the other hand, the proposed method tries to inactivate the adversarial perturbations by projecting the adversarial samples into the immune space where the recognition model is robust to perturbations.
The gradient direction can be effectively generated by reparameterization.

\item The degree of randomness is different.
In the proposed method, the randomness derives from the sampling from Bernoulli distributions, which means the random variables are binary.
Meanwhile, most of the sampling probabilities are close to 0 or 1, as shown in Fig.~\ref{fig:ValProb} (b), which further reduces the degree of randomness.
On the other hand, in most of the stochastic gradient methods, the random variables are continuous or approximately continuous.
And they usually deliberately increase the randomness degree to improve the performance.
If we measure the randomness by information entropy, the entropy of the stochastic gradient methods would be much bigger than the proposed methods.

\end{itemize}

Besides, to verify the difference between obfuscated gradient methods and the proposed method, two experimental analyses are provided:


\subsubsection{Apply the Five Conditions of Obfuscated Gradient Methods}

Athalye et al.~\cite{athalye2018obfuscated} summarized five conditions to identify obfuscated gradient methods.
To verify the difference between the proposed method and obfuscated gradient method, all the five conditions are applied to the proposed method:

(i): One-step attacks should not perform better than iterative attacks  for non-obfuscated gradient methods.

For obfuscated gradient methods, the one-step attacks may perform better than iterative attacks since the gradient direction is disturbed.
I-FGSM~\cite{Kurakin2017Adversarial}, which is an iterative attack method based on FGSM, is adopted in this test.
Only online defense protocols are adopted.
The results on LFW are shown in Tab.~\ref{tab:IFGSM}.
Obviously, the iterative attack I-FGSM outperforms the one-step attack FGSM.

\begin{table}[h]
\begin{center}
\setlength{\abovecaptionskip}{0pt}
\setlength{\belowcaptionskip}{0pt}
\caption{The iterative attack I-FGSM outperforms the one-step attack FGSM.}
\label{tab:IFGSM}
\setlength{\tabcolsep}{4mm}
{
\begin{tabular}{c|c|c}
\hline
& \bf{FGSM}-online & \bf{I-FGSM}-online  \\
\hline
\bf{EER} & 14.51\% & 26.73\% \\
\hline
\end{tabular}}
\end{center}
\end{table}

(ii): Black-box attacks should not perform better than white-box attacks for non-obfuscated gradient methods.

For obfuscated gradient methods, the transfer-based black-box attacks may perform better than white-box attacks.
In this test, DI$^2$-FGSM~\cite{xie2019improving}, which is a transfer-based black-box attack method, is adopted to verify the proposed method on LFW.
DI$^2$-FGSM is performed on the recognition backbone without defense, and the generated adversarial samples are transferred to attack the proposed method.
The results are shown in Tab.~\ref{tab:DIFGSM}.
The EER of the transfer-based black-box attack is just slightly higher than FGSM-offline, whose target is also the recognition backbone without defense.
The FGSM-online obviously outperforms the transfer-based black-box attack.
The results verify that the proposed method is different from the obfuscated gradient methods.

\begin{table}[h]
\begin{center}
\setlength{\abovecaptionskip}{0pt}
\setlength{\belowcaptionskip}{0pt}
\caption{Perform the transfer-based black-box attack. The FGSM-online obviously outperforms the transfer-based black-box attack, which verifies the differences between the proposed method and obfuscated gradient methods.}
\label{tab:DIFGSM}
\setlength{\tabcolsep}{4mm}
{
\begin{tabular}{c|c|c|c}
\hline
& \bf{FGSM}-offline & \bf{FGSM}-online & \bf{DI$^2$-FGSM} \\
\hline
\bf{EER} & 6.15\% & 14.51\% & 6.81\% \\
\hline
\end{tabular}}
\end{center}
\end{table}

(iii)\&(iv): Unbounded attacks can reach 100\% success, and increasing distortion bound should increase success for non-obfuscated gradient methods.

For obfuscated gradient methods, increasing distortion bound does not increase the success rate of attack, and unbounded attacks do not reach 100\% success.
We increase the distortion bound of FGSM to test the proposed method.
In this test, the threshold of similarity is set to 0.2 (the possible range is $[-1, 1]$), which means that image pairs with a similarity higher than 0.2 are predicted as positive pairs.
Only online defense protocols are adopted.
All of the positive pairs in Section~\ref{ExpIC} are adopted, and the error rate of prediction is equivalent to the success rate of the attack in this setting.
The results are shown in Tab.~\ref{tab:increas_FGSM}.
The error rate increases along with the distortion bound, and reaches 100\% when the distortion bound is 0.75.

\begin{table}[h]
\begin{center}
\setlength{\abovecaptionskip}{0pt}
\setlength{\belowcaptionskip}{0pt}
\caption{The error rate increases along with the distortion bound, and reaches 100\% when distortion bound is 0.75.}
\label{tab:increas_FGSM}
\setlength{\tabcolsep}{4mm}
{
\begin{tabular}{c|c|c|c|c}
\hline
\bf{Bound} & 0.04        & 0.06       & 0.08       & 0.1          \\
\hline
\bf{EER}    & 13.74\% & 30.63\% & 44.63\% & 55.65\% \\
\hline
\hline
\bf{Bound} & 0.2         & 0.3         & 0.5          & 0.75 \\
\hline
\bf{EER}    & 93.08\% & 98.57\% & 99.42\% & 100\%\\
\hline

\end{tabular}}
\end{center}
\end{table}

(v): Random sampling should not find adversarial examples for non-obfuscated gradient methods.

Brute-force random search should not find adversarial examples when gradient-based attacks do not.
However, for obfuscated gradient methods, brute-force random search (e.g., randomly sampling $10^5$ or more points) would find adversarial samples within the distortion bound when gradient-based attacks do not.
In this test, the distortion bound of PGD-online is set to 0.01, and the threshold of similarity is set to 0 (the possible range is $[-1, 1]$, the success condition is stricter so that PGD cannot succeed on most pairs) when the attack success rate on the positive pairs of LFW is 3.10\%.
The success rate of brute-force random search under the same protocol (sampling $10^5$ points for each pair)  is 0.27\%.
The brute-force search strategy only succeeds in the pairs whose similarities are nearby 0 or lower than 0 even without attacks.
No adversarial samples can be found by brute-force search if PGD-online cannot.


\subsubsection{Perform BPDA Attack and High-iteration PGD}

Athalye et al.~\cite{athalye2018obfuscated} propose an effective way to circumvent obfuscated gradient methods called Backward Pass Differentiable Approximation (BPDA).
BPDA finds an approximation of the non-differentiable layer during the backward pass to estimate the gradient.
To apply the BPDA attack, we replace the sampling operation in the proposed method with a binarization operation during the backward pass: the output of the agent is directly binarized according to a threshold of 0.5.
The components higher than 0.5 is set to 1, and others are set to 0.
It is differentiable and has no more randomness during the backward pass.
Meanwhile, high-iteration (100) PGD is also adopted in this experiment.
Only online defense protocols are adopted.

The results are shown in Tab.~\ref{tab:BPDA}.
The EERs of BPDA attack are just slightly higher than the original gradient-based attacks.

\begin{table}[h]
\begin{center}
\setlength{\abovecaptionskip}{0pt}
\setlength{\belowcaptionskip}{0pt}
\caption{``Orig" denotes the original attacks. ``BPDA" denotes the BPDA attacks.}
\label{tab:BPDA}
\setlength{\tabcolsep}{4mm}
{
\begin{tabular}{c|c|c|c}
\hline
& \bf{FGSM} & \bf{PGD} & \bf{PGD-100} \\
\hline
\bf{Orig} & 14.51\% & 63.32\% & 88.37\% \\
\hline
\bf{BPDA} &   15.78\%     & 68.19\% & - \\
\hline

\end{tabular}}
\end{center}
\vspace{-1cm}
\end{table}


%


\begin{table*}[h]
\begin{center}
\setlength{\abovecaptionskip}{0pt}
\setlength{\belowcaptionskip}{0pt}
\caption{Ablation studies on LFW}
\label{tab:AblGen}
\setlength{\tabcolsep}{5mm}
{
\begin{tabular}{c|c|c|c|c|c}
\hline
\multicolumn{2}{c|}{\bf{Attack Method}}  & \bf{Clean} & \bf{FGSM} & \bf{DeepFool} & \bf{PGD} \\
\hline
\multirow{8}*{Offline}& No Defense (ArcFace)  & 0.44\%        & 41.97\%     &89.49\%      & 99.71\%  \\
\cline{2-6}
& No Defense (LightCNN)                                    & 1.08\%        & 51.06\%     &91.29\%       & 99.73\%  \\
\cline{2-6}
& PCA-200                                                               & 9.06\%        & 9.42\%      &12.35\%        & \bf{22.93}\%  \\
\cline{2-6}
 & PCA-2000                                                            & \bf{0.53}\% & 23.39\%    &72.85\%       & 93.45\%  \\
 \cline{2-6}
 & DAE Only                                                           & 2.18\%        & 12.56\%       &25.95\%         & 54.07\%  \\
\cline{2-6}
 & Ours-LightCNN                                                 & 4.11\%        & 11.55\%       &12.16\%         & 29.98\%  \\
 \cline{2-6}
  & Ours-without DAE                                           & 1.94\%        & \bf{6.09}\%  &9.14\%         & 30.17\%  \\
 \cline{2-6}
 & Ours                                                                    & 1.95\%        & 6.15\%         &\bf{7.86}\% &  29.77\%  \\

\hline
\hline

\multirow{8}*{Online} & No Defense (ArcFace)  & 0.44\%         & 41.97\%         &89.49\%       & 99.71\%  \\
\cline{2-6}
& No Defense (LightCNN)                                    & 1.08\%          & 51.06\%         &91.29\%       & 99.73\%  \\
\cline{2-6}
& PCA-200                                                               & 9.06\%.         & 43.03\%           &49.93\%         & 93.18\%  \\
\cline{2-6}
 & PCA-2000                                                            & \bf{0.53}\%  & 40.09\%           &82.87\%         & 99.03\%  \\
\cline{2-6}
& DAE Only                                                           & 2.18\%        & 75.69\%       &95.23\%         & 99.09\%  \\
\cline{2-6}
 & Ours-LightCNN                                                 & 4.11\%          & 23.52\%           &40.58\%         & 67.16\%  \\
 \cline{2-6}
   & Ours-without DAE                                          & 1.94\%           & 14.73\%.          &\bf{33.69}\%  & 63.61\%  \\
 \cline{2-6}
 & Ours                                                                   & 1.95\%            & \bf{14.51}\% & 34.66\% & \bf{63.32}\%  \\

\hline
\end{tabular}}
\end{center}
\vspace{-0.5cm}
\end{table*}

\subsection{Ablation Study}
\label{ExpAbl}
In this subsection, we provide further investigations of the proposed method in three aspects: (1) Is the proposed learnable PCA better than classical PCA for adversarial defense? (2) Is the proposed method a plug-and-play framework for face recognition models? (3) What is the impact of DAE on the performance?
The experiments are launched on LFW, and the evaluation protocol is the same as Section ~\ref{ExpIC}.

\subsubsection{Comparison with PCA}
The proposed learnable PCA selects eigenvectors for each sample by the trained agent. On the other hand, classical PCA selects eigenvectors according to eigenvalues, and the selection is the same for all samples.
For comparison with the learnable PCA, the first 200 and 2000 eigenvectors are selected separately for classical PCA reconstruction.
The results are shown in Tab.~\ref{tab:AblGen}.

It is observed from Tab.~\ref{tab:AblGen} that for classical PCA, the selection of eigenvectors is a trade-off between robustness against adversarial perturbations and recognition performance.  A more precise reconstruction of the input image is obtained when more eigenvectors are selected.
However, there is less robustness against attack because the adversarial perturbations are also reconstructed.
On the other hand, the proposed learnable PCA significantly improves the robustness against attack at a much lower cost of recognition performance.

The eigenvectors with maximal eigenvalues are optimal when all of the samples are taken into consideration. However, for a single image, eigenvectors with maximal eigenvalues can be suboptimal.
By contrast, learnable PCA finds the appropriate eigenvectors for every single sample. Eigenvalues of LFW are shown in Fig.~\ref{fig:ValProb} (a). The probabilities of the eigenvectors of a specific sample are shown in Fig.~\ref{fig:ValProb} (b). These two figures show the differences in the selection manners. The eigenvectors with low eigenvalues may have high probabilities to be selected for a special sample and vice versa.


\subsubsection{Combination with Different Face Recognition Models}
Two popular face recognition models are taken into consideration in this experiment: ArcFace~\cite{Deng2018ArcFace} and LightCNN~\cite{wu2018light}.
The results are shown in Tab.~\ref{tab:AblGen}. The results of ArcFace are marked as ``Ours'', and the results of LightCNN are marked as ``Ours-LightCNN''.

The results in Tab.~\ref{tab:AblGen} show that the proposed method significantly improves the robustness against adversarial attacks for both ArcFace~\cite{Deng2018ArcFace} and LightCNN~\cite{wu2018light}.
It is revealed that the proposed adversarial defense method is a plug-and-play method and is independent of the recognition models. This is an overwhelming advantage in practice because it means that this method can be easily applied to existing face recognition systems.

\subsubsection{Impact of DAE}
The denoising autoencoder in the proposed method is mainly employed to accelerate the convergence of the training process as mentioned above in Section~\ref{MetLeaPCA}. The impact of the denoising autoencoder on the defense performance is explored in this experiment.

The results without DAE are shown in Tab.~\ref{tab:AblGen} (``Ours-without DAE''). We observe that the performance characteristics are very close to those of the original method and are even slightly better in some cases. 
On the other hand, the performance of ``DAE Only" declines sharply compared to the original framework, especially in the online defense protocol.
 These results indicate that the denoising autoencoder has only a marginal effect on the performance.


\section{Conclusion}
\label{Conclusion}

Although deep learning-based face recognition models have made significant progress, they lack defense mechanisms against adversarial attacks.
In this paper, we explore the effects of perturbations in different subspaces on the similarity metrics of face recognition.
It is found that the subspace of perturbation is a key factor in the impact of the attack on the recognition models.
Based on this discovery, we propose a novel adversarial defense method to stimulate the inherent robustness against perturbations of deep models to resist adversarial attacks.
The learnable PCA is proposed to piecewise-linearly estimate the immune space, and the adversarial perturbations are inactivated by projecting the input image into this subspace.
The experimental results demonstrate the superior generalization capacity of the proposed method to different kinds of adversarial perturbations.
The proposed approach is a plug-and-play method that is successfully applied to four commercial APIs without extra training and significantly improves the robustness against unseen attacks.



%



\section*{Acknowledgment}

The authors would like to thank the associate editor and the reviewers for their valuable comments and advices.
This work is supported by the National Natural Science Foundation Of China (Grant Nos.  U1836217, 62006225, 61622310, 62071468) and the Strategic Priority Research Program of Chinese Academy of Sciences (Grant No. XDA27040700), sponsored by CAAI-Huawei Mindspore Open Fund.

\ifCLASSOPTIONcaptionsoff
  \newpage
\fi

\end{document}